\documentclass[11pt]{article}
\usepackage{setspace}
\usepackage{amsmath,amssymb}
\usepackage{amsthm}
\usepackage{algorithm, algpseudocode}
\usepackage{url}
\usepackage{fullpage}
\usepackage{color}
\usepackage{graphicx}

\newtheorem{claim}{Claim}[section]
\newtheorem{lemma}[claim]{Lemma}

\newtheorem{theorem}{Theorem}
\newtheorem{proposition}[claim]{Proposition}

\newtheorem{definition}[claim]{Definition}

\theoremstyle{definition}

\def\Binom{{\rm Binom}}
\def\ex{\mbox{\rm\tiny ex}}
\def\fr{\mbox{\rm\tiny fr}}
\def\pl{\mbox{\rm\tiny pl}}

\def\frl{\mbox{\rm fr}}
\def\pll{\mbox{\rm pl}}
\def\sTV{\mbox{\rm\tiny TV}}

\def\var{{\rm var}}
\def\cC{{\cal C}}
\def\cG{{\cal G}}
\def\Loc{{\sf Loc}}
\def\hS{\widehat{S}}
\def\de{{\rm d}}
\def\Psucc{{\rm P}_{\rm succ}}
\def\ed{\stackrel{{\rm d}}{=}}

\def\P{{\rm P}}

\def\<{\langle}
\def\>{\rangle}
\def\prob{{\mathbb P}}

\def\naturals{{\mathbb N}}
\def\E{{\mathbb E}} 
\def\Er{{\rm E}} 
\def\Var{{\sf{Var}}}

\def\reals{\mathbb{R}}
\def\sT{{\sf T}}
\def\E{\mathbb{E}}
\def\normal{{\sf N}}
\def\F{{\sf F}}

\def\cF{{\cal F}}

\def\eps{\varepsilon}

\def\ind{{\mathbb{I}}}

\def\Ball{{\sf B}}
\def\bBall{\overline{\sf B}}

\def\Ber{{\rm Bernoulli}}

\def\bone{{\mathbf 1}}

\def\ox{\overline{x}}

\def\Pv{\Psi^{{\rm v}}}
\def\Pe{\Psi^{{\rm e}}}
\def\Free{\Psi}
\def\free{\psi}
\def\omu{{\overline{\mu}}}

\def\tp{\tilde{p}}

\def\Info{{\rm I}}
\def\Poisson{{\rm Poisson}}
\def\Bernoulli{{\rm Bernoulli}}

\author{
Andrea~Montanari\footnote{Department of Electrical
    Engineering and Department of Statistics, Stanford University}}
\title{Finding One Community in a Sparse  Graph}
\begin{document}
\maketitle

\begin{abstract}
We consider a random sparse graph with bounded average degree, in which
a subset of vertices has higher connectivity 
than the background. In particular, the average degree inside
this subset of vertices is larger than outside (but still bounded). Given a
realization of such graph, we aim at  identifying the
hidden subset of vertices.  This can be regarded as a model for the
problem of finding a tightly knitted community in a social network, or 
a cluster in a relational dataset.

In this paper we present two sets of contributions: $(i)$ We use
the cavity method from spin glass theory to derive an exact phase
diagram for the reconstruction problem. In particular, as the
difference in edge probability increases, the problem undergoes two
phase transitions, a static phase transition and a dynamic one. $(ii)$
We establish rigorous bounds on the dynamic phase transition and prove
that, above a certain threshold, a local algorithm  (belief
propagation) correctly identify most of the hidden set. Below the
same threshold \emph{no local algorithm} can achieve this
goal. However, in this regime the subset can be identified by
exhaustive search.

For small hidden sets and large average degree, the phase transition 
for local algorithms takes an intriguingly simple form. Local
algorithms succeed with high probability  for $\deg_{\rm in} - \deg_{\rm out} >
\sqrt{\deg_{\rm out}/e}$ and fail for $\deg_{\rm in} - \deg_{\rm out} <
\sqrt{\deg_{\rm out}/e}$ (with $\deg_{\rm in}$, $\deg_{\rm out}$ the
average degrees inside and outside the community).
We argue that spectral algorithms are also ineffective in the latter regime.
 It is an open
problem whether any polynomial time algorithms might succeed 
for $\deg_{\rm in} - \deg_{\rm out} <
\sqrt{\deg_{\rm out}/e}$. 
\end{abstract}

\section{Introduction}

\subsection{Motivation}

The problem of finding a highly connected subset of vertices in a large
graph arises in a number of applications across science and
engineering.  Within social network analysis, a highly connected
subset of nodes is interpreted as a community \cite{fortunato2010community}. Many approaches
to data clustering and dimensionality reduction construct a `similarity
graph' over the data points. A highly connected subgraph corresponds
to a cluster of similar data points \cite{von2007tutorial}.

A closely related problem arises in the analysis of matrix data,
e.g. in microarray data analysis. In this context, researchers are
often interested in a submatrix whose entries have an average
value larger (or lower)   than the rest
\cite{shabalin2009finding}.
Such an anomalous submatrix is interpreted as evidence of association between gene expression levels 
and phenotypes (e.g. medical conditions).
If we consider the graph adjacency matrix, a highly connected subset
of vertices corresponds indeed to a principal submatrix with average
value larger than the background.

The special case of finding a completely connected subset of vertices
(a clique) in a graph has been intensely studied within theoretical
computer science. Assuming P$\neq$NP,  the largest clique in a
graph cannot be found in polynomial time. Even  a very rough
approximation to its size is hard to find
\cite{hastad1996clique,khot2001improved}. In particular, it is hard to
detect the presence of a clique of size $N^{1-\eps}$ in a graph with
$N$ vertices.

Such hardness results motivated the study of random instances. In
particular,
 the so-called `planted clique' or `hidden clique problem'
 \cite{jerrum1992large} requires to find a clique of size $k$  that is
 added (planted) in a
 random graph with edge density $1/2$. More precisely, for a subset of
 vertices $S\subseteq [N]$, all edges $(i,j)$, with $\{i,j\}\subseteq S$
 are present. All other edges are present independently with
 probability $1/2$.  Such a clique can be found reliably by exhaustive
 search as soon as  $k\ge 2(1+\eps)\log_2N$ \cite{grimmett1975colouring}. However, despite many
 efforts, no algorithm is
 known that achieves this goal for $k\ll \sqrt{N}$ \cite{alon1998finding,feige2000finding,dekel2014finding}.
In other words, the problem of finding cliques of size $2\log_2 N\ll
k\ll \sqrt{N}$ is solvable, but possibly hard. Proving that indeed it
is computationally hard to  find cliques in this regime is an
outstanding problem in theoretical computer science.

For  general polynomial algorithms, it  is known since \cite{alon1998finding}
  that a clique of size $\delta\sqrt{N}$ can be found in time
  $N^{O(\log(1/\delta))}$ for any $\delta>0$ fixed. Hence, if we allow
  any time complexity polynomial in $N$, then the question is whether
the planted clique can be found for $k = o(\sqrt{N})$.

A more stringent computational constraint  requires that the 
clique is found in nearly-linear time, i.e. in time of order
$O(N^2 (\log N)^c)$. Note that the number of bits required to encode
an instance of the problem is of order $N^2$, so $N^2(\log N)^c$ is a
logarithmic multiple of the time required to read an instance. 
Dekel, Gurel-Gurevittch and Peres \cite{dekel2014finding} developed a
linear-time algorithm (i.e. with complexity $O(N^2)$) that finds the
hidden clique with high probability, provided $k>1.261\sqrt{N}$.
In \cite{deshpande2013finding} it was proved that, if
$k>(1+\eps)\sqrt{N/e}$, 
then there exists a message passing algorithm that finds
  with high probability the clique with $O(N^2\log N)$ operations.
The same paper
provided evidence that a certain class of `local' algorithms
fails at the same threshold.  Among other motivations, the present
paper generalizes and supports the existence of a fundamental
threshold for local algorithms --at least in the sparse graph setting.

\subsection{Rigorous contributions}

In the present paper, we consider the problem of finding
a highly connected subset of vertices in a \emph{sparse graph},
i.e. in a  graph with \emph{bounded average degree}.  
In this case, the hidden set size must scale linearly with $N$ to
obtain a non-trivial behavior.
Somewhat surprisingly, 
we find that the phase transition `at $1/\sqrt{e}$' leaves a trace
also in the sparse regime.

More precisely, we consider a random graph generated as follows.
We select a subset of vertices $S$ of size $\kappa N$, uniformly at random
given its size. We connect any two vertices in the set independently
with probability $a/N$. Any other edge is added independently with
probability $b/N$, $b<a$.
The problem distribution is therefore parametrized by
$a,b,\kappa\in\reals$
and we will be therefore interested in the limit $N\to\infty$, with
$a,b,\kappa$ fixed. A more intuitive parametrization is obtained by 
replacing $a,b$  with the average degrees for vertices $i\in S$, and
$i\not \in S$ denoted, respectively, by $\deg_{\rm in}$, $\deg_{\rm out}$

Our main rigorous result is a sharp phase transition in the following
double asymptotics:
\begin{itemize}
\item First $N\to\infty$. This corresponds to considering large graphs.
\item Then $\kappa\to 0$ and $\deg_{\rm in},\deg_{\rm out}\to\infty$.
  This corresponds to focusing on small hidden sets, but still linear in
  $N$. The requirement $\deg_{\rm in},\deg_{\rm out}\to\infty$ is
  a necessary consequence of $\kappa\to 0$: it can be shown that otherwise the hidden set cannot
  possibly be detected.
\end{itemize}

Our main rigorous result (Theorem \ref{thm:Main}) establishes that, in the above double asymptotics, 
a phase transition takes place for local algorithms at
\begin{align}
\deg_{\rm in}-\deg_{\rm out} = \sqrt{\frac{\deg_{\rm out}}{e}}\, . \label{eq:BasicCondition}
\end{align}
Namely, we  consider the problem of testing whether a vertex $i$ is in $S$
or not. We say that such a test is reliable if, in the above limit,
the fraction of incorrectly estimated vertices vanishes in
expectation. Then:
\begin{itemize}
\item For $\deg_{\rm in}-\deg_{\rm out} >(1+\eps) \sqrt{\deg_{\rm out}/e}$, a
  local algorithm can estimate reliably $S$ in time of the order of
  the number of edges. This is achieved for instance, by the
  belief propagation algorithm. 
\item For $\deg_{\rm in}-\deg_{\rm out}< (1-\eps)\sqrt{\deg_{\rm out}/e}$, no
  local algorithm can reliably reconstruct $S$.
\end{itemize}
Analogously to the classical hidden clique problem, there is a large gap
between what can be achieved by local algorithms, and optimal
estimation with unbounded computational resources.
Proposition \ref{propo:Exhaustive} estabilishes that  exhaustive
search will find $S$ in exponential time, as soon as $\deg_{\rm in}-\deg_{\rm out} >\eps \sqrt{\deg_{\rm out}/e}$ for
some positive $\eps$ (in the same double limit).

Note that, in both cases, a small fraction of the vertices in
  $S$ remains undetected because of the graph sparsity, for any
  $\deg_{\rm in}<\infty$. In
  particular,  the number of vertices of degree $0$ is linear in $N$,
  and such nodes cannot be identified.

Let us finally mention the degree of a vertex $i$ is a Poisson with mean $\deg_{\rm in}$
if $i\in S$ and mean $\deg_{\rm out}$ if $i\not \in S$. Hence, the
degree standard deviation (outside $S$) is $\sqrt{\deg_{\rm
    out}}$. Therefore, the
ratio $(\deg_{\rm in}-\deg_{\rm out})/\sqrt{\deg_{\rm out}}$ is the
difference in mean degree divided by the standard deviation, and has the
natural interpretation of a `signal-to-noise ratio.'

\subsection{Non-rigorous contributions}

While our rigorous analysis focuses on the limit $\kappa\to 0$ and
$\deg_{\rm in},\deg_{\rm out}\to\infty$
(after $N\to\infty$), we will use the cavity method from spin glass
theory to investigate the model behavior for \emph{arbitrary} $\deg_{\rm
  in},\deg_{\rm out},\kappa$ (in the $N\to\infty$ limit) or,
equivalently,
arbitrary $a,b,\kappa$.

We will use two approaches to obtain concrete predictions from the
cavity method:
\begin{itemize}
\item For general $a,b$ (bounded degree), we derive the cavity
predictions for local quantities, as well as for the free energy
density. We show that this indeed coincide (up to a shift) with the
mutual information per variable between the hidden set $S$ and the
observed graph $G$.

 We use the `population dynamics' (or `sampled
density evolution') algorithm
\cite{mezard2001bethe,RiU08,MezardMontanari} to solve numerically the
cavity equations.
\item We then consider the limit of large $a,b$ (large degree) for
  arbitrary  $\kappa$. In order to obtain a non-trivial limit, the
  signal-to-noise ratio $\lambda= \kappa^2(a-b)^2/[(1-\kappa)b]$ is
  kept fixed in this limit, together with $\kappa\in [0,1]$.

The cavity equations simplify in this limit (the cavity field
distributions become Gaussian), and we can derive an exact phase
diagram, without recourse to intensive numerical methods, cf. Figure \ref{fig:PhaseDiagLargeDeg}.
\end{itemize}
This two approaches are complementary in that the large-degree 
asymptotics yields closed-form expressions. The qualitative features of
the resulting phase diagram should remain unchanged at moderately
small values of $a,b$. Our population dynamics analysis confirms this.

\vspace{0.5cm}

As already mentioned, one of the motivations for the present work was
to better understand the computational phase transition discovered in \cite{deshpande2013finding}
for the classical hidden clique problem.
For background edge density $1/2$, this takes place when the size of the hidden clique is $k\approx\sqrt{N/e}$.
This phase transition can indeed  be formally recovered as a dense limit 
of the results presented in this paper. 

More precisely, the phase transition for hidden cliques 
\cite{deshpande2013finding} is captured by
Eq.~(\ref{eq:BasicCondition}),
once we rewrite the latter in terms $\var_{\rm
  out}$,  the variance of the degrees of nodes $i\not \in S$. 
 In the sparse regime, the degree is
approximately Poisson distributed, and hence $\var_{\rm out}\approx
\deg_{\rm out}$. 
Therefore Eq.~(\ref{eq:BasicCondition}) can be  rewritten as $(\deg_{\rm in}-\deg_{\rm out})/\sqrt{\var_{\rm
    out}} = 1/\sqrt{e}$
For the classical (dense) hidden clique
problem, we have  $\deg_{\rm out}= (N-1)/2$, $\deg_{\rm in} =
(N+k-2)/2$ and $\var_{\rm out} = (N-1)/4$, and hence we recover the
condition $k\approx \sqrt{N/e}$.

From a different perspective, the present work offers a 
statistical mechanics interpretation of the phase transitions in the
 hidden clique problem.
 Namely, the latter can be formally  recovered as the $\kappa\to 0$
 limit  of the phase diagram in Figure
\ref{fig:PhaseDiagLargeDeg} below. In particular, the computational
phase transition at $k = \sqrt{N/e}$ corresponds to a dynamical
phase transition (a spinodal point) in the underlying statistical
mechanics model.

\subsection{Paper outline}

The rest of the paper is organized as follows. 
In the next section we define formally our model and some related
notations. Section \ref{sec:Cavity} derives the phase diagram using the
cavity method. In particular, we show that the model undergoes two
phase transitions as the signal-to-noise ratio increases (for $k/N$ small enough): a
static phase transition and a dynamic one,
The two phase transitions are well separated.
Section \ref{sec:Rigorous} presents rigorous
bounds on the behavior of local algorithms and exhaustive search, that
match the above phase transitions for small $k/N$. This section is
self-contained and the interested reader can  move directly to it,
after the model definition (some useful, but elementary results are
presented in Section \ref{sec:TreeInterpretation}).
Proofs are deferred to the appendix.
Finally, Section \ref{sec:Related} positions our results in the
context of recent literature.

Several research communities have been working on closely 
related problems: statistical physics, theoretical computer science,
machine learning, statistics, information theory. We tried to write a paper that could be 
accessible to researchers with different backgrounds, both in terms of
tools and of language. We apologize for any redundancy that might have
followed from this approach.

\subsubsection*{Notations}
 
We use $[\ell]= \{1,\dots,\ell\}$ to denote the set
of first $\ell$ integers,  and $|A|$ to denote the size (cardinality)
of set $A$.)
For a set $V$, we write $(i,j)\subseteq V$ to indicate that $(i,j)$ runs
over all unordered pairs of distinct elements in $V$. For instance,
for a symmetric function $F(i,j) = F(j,i)$, we have
\begin{align}
\prod_{(i,j)\subseteq [N]} F(i,j) \equiv \prod_{1\le i<j\le N} F(i,j)\, .
\end{align}
If instead $E$ is a set of edges over the vertex set $V$ (unordered
pairs with elements in $V$) we write $(i,j)\in E$ to denote elements
of $E$.

We use $\normal(\mu,\sigma^2)$ to denote the Gaussian distribution
with mean $\mu$ and variance $\sigma^2$. Other classical probability
distributions are denoted in a way that should be self-explanatory 
(Bernoulli$(p)$, Poisson$(c)$, and so on).
\section{Model definition}
\label{sec:Model}

We consider a random graph $G_N = (V_N, E_N)$ with vertex set
$V_N = [N] \equiv \{1,\dots,N\}$ and random edges generated as follows. A set $S\subseteq
V_N$ is chosen at random. Introducing the indicator variables
\begin{align}
x_i = \begin{cases}
1& \;\;\mbox{if $i\in S$,}\\
0& \;\;\mbox{otherwise,}\
\end{cases}
\end{align}
we let $x_i\in\{0,1\}$ independently with 
\begin{align}
\prob\big(x_i = 1\big) = \kappa\, . \label{eq:KappaDef}
\end{align}
In particular $|S|$ is a binomial random variable, and is tightly
concentrated around its mean $\E|S| = \kappa N$.
Edges are independent given $S$, with the following probability for
$i,j\in V_N$ distinct: 
\begin{align}
\prob\big\{(i,j)\in E_N|S\big\} = \begin{cases}
a/N & \;\; \mbox{if $\{i,j\}\subseteq S$,}\\
b/N & \;\; \mbox{otherwise.}
\end{cases}
\end{align}

We let $x=(x_1,\dots,x_N)$ denote the vector identifying $S$.
By using Bayes theorem, the conditional distribution of $x$ given $G$
is easily written
\begin{align}
\tp_G(x) \equiv \prob(x|G) = \frac{1}{\widetilde{Z}(G)} \prod_{i\in
  [N]}\Big(\frac{\kappa}{1-\kappa}\Big)^{x_i}
\prod_{(i,j)\subseteq [N]}
\Big(\frac{1-a/N}{1-b/N}\Big)^{x_ix_j}\prod_{(i,j)\in E} 
\rho_N^{x_ix_j}\, ,\label{eq:ExactPosterior}
\end{align}
where $\rho_N\equiv (a/b)\big((1-b/N)/(1-a/N) \big)$.

We next replace the last probability distribution with
one that is equivalent as $N\to\infty$, and slightly more convenient for 
 the cavity calculations of the next section. (These simplifications
 will not be used to prove the rigorous bounds in Section
 \ref{sec:Rigorous}.)
We first note that, as $N\to\infty$, we have $\rho_N\to\rho$ with
\begin{align}
\rho \equiv \frac{a}{b} \, . \label{eq:RhoDef}
\end{align}
Next, letting $|x| \equiv \sum_{i=1}^Nx_i$, we can rewrite the second product in
Eq.~(\ref{eq:ExactPosterior}) as 
\begin{align}
\prod_{(i,j)\subseteq [N]} &
\Big(\frac{1-a/N}{1-b/N}\Big)^{x_ix_j} =
\Big(\frac{1-a/N}{1-b/N}\Big)^{\binom{|x|}{2}}
 = C\, 
\Big(\frac{1-a/N}{1-b/N}\Big)^{(2\kappa N-1)(|x|-\kappa N)/2} \cdot
 \Big(\frac{1-a/N}{1-b/N}\Big)^{\frac{1}{2}(|x|-\kappa N)^2}\label{eq:SimplificationGlobal}
\end{align}
with
\begin{align}
C = \Big(\frac{1-a/N}{1-b/N}\Big)^{\frac{1}{2}\kappa N(\kappa N-1)}\,
,
\end{align}
a constant independent of $x$.
Notice that $|x|\sim \Binom(N,\kappa)$ is tightly concentrated around
$\E\{|x| \}= \kappa N$. In particular $|x| = \kappa N+ O(\sqrt{N})$
with high probability, and therefore the last term in
Eq.~(\ref{eq:SimplificationGlobal})
is of order $\Theta(1)$. We will neglect it, thus
obtaining
\begin{align}
\prod_{(i,j)\subseteq [N]} &
\Big(\frac{1-a/N}{1-b/N}\Big)^{x_ix_j}  \approx   
 C'\, 
\Big(\frac{1-a/N}{1-b/N}\Big)^{\kappa N(|x|-\kappa N)} \approx 
 C'' \, e^{-\kappa(a-b)|x|}\, .
\end{align}
The error incurred by neglecting the  last term in
Eq.~(\ref{eq:SimplificationGlobal}) can be corrected by 
considering the following approximate
conditional distribution of $x$ given the graph $G$ 
\begin{align}
p_G(x) = \frac{1}{Z(G)}\;\prod_{(i,j)\in E}\rho^{x_ix_j}
\prod_{i\in V}\gamma^{x_i}\, \ind\Big(\sum_{i\in V_N}x_i=\kappa N\Big) \label{eq:ApproxModel}
\end{align}
where $\ind(A)$ is the indicator function on condition $A$ and 
\begin{align}
\gamma & \equiv e^{-\kappa(a-b)} \Big(\frac{\kappa}{1-\kappa}\Big) \, .
\end{align}
 Note that we multiplied $\tp_G(\,\dot\,)$ by the indicator function
 $\ind\Big(\sum_{i\in V_N}x_i=\kappa N\Big)$. 
This can be interpreted as replacing the i.i.d. Bernoulli distribution
(\ref{eq:KappaDef}) with the uniform distribution over $S$ with
$|S|=\kappa N$, which is immaterial as long as local properties of
$\tp_G(x)$ are considered. 

In the following, we shall compare different reconstruction methods.
Any such method corresponds to a function $T_i(G)\in\{0,1\}$ of vertex $i$
and graph $G$, with the interpretation
\begin{align}
T_i(G) = \begin{cases}
1 & \mbox{ if $i$ is estimated to be in $S$},\\
0 & \mbox{ if $i$ is estimated not to be in $S$}.
\end{cases}
\end{align}
We characterize such a test through its rescaled success probability
\begin{align}
\Psucc^{(N)}(T) = \prob\big(T_i(G) = 1\big| i\in S\big) + \prob\big(T_i(G) =
0\big|i\not\in S\big)-1 \, . \label{eq:PsuccT}
\end{align}
Note that a trivial test (assigning $T_i(G)\in\{0,1\}$  at random
independently of $G$) achieves $\Psucc^{(N)}(T) = 0$, while a perfect
test has $\Psucc^{(N)}(T) = 1$.  We shall often omit the arguments $T$, $n$
from $\Psucc^{(N)}(T) $ in the following.

We note in passing that the optimal estimator with respect to the
metric (\ref{eq:PsuccT})
is the maximum-likelihood estimator
\begin{align}
T_i^{\rm opt}(G) = \begin{cases}
1 & \mbox{ if $\prob(G|i\in S)\ge \prob(G|i\not \in
S)$,}\\
0 & \mbox{ if $\prob(G|i\in S)< \prob(G|i\not \in
S)$.}
\end{cases}
\end{align}
Namely, for any other estimator $T$, we have $\Psucc^{(N)}(T) \le
\Psucc^{(N)}(T^{\rm opt})$ (see, for instance, the textbook
\cite{Lehmann1998} for a proof of this fact).
The resulting success probability coincides with the total variation
distance between the conditional distribution of $G$ given the two
hypotheses $i\in S$ and $i\not\in S$. Recall that, given two
probability measures $p, q$ on the same finite space $\Omega$, their
total variation distance is defined as $\|p(\,\cdot\,)-q(\,\cdot\,)\|_{\sTV}\equiv
(1/2)\sum_{\omega\in \Omega}|p(\omega)-q(\omega)|$.
Then we have
\begin{align}
\Psucc^{(N)}(T) \le \Psucc^{(N)}(T^{\rm opt}) = \|\prob(G\in \cdot
|i\in S)\ge \prob(G\in\cdot |i\not \in S)\|_{\sTV}\, .
\end{align}
%

%
%
\section{Phase transitions via cavity method}
\label{sec:Cavity}

In this section we use the cavity method to derive an exact phase
diagram of the model. It is convenient to introduce the following
signal-to-noise-ratio parameter:
\begin{align}
\lambda \equiv \frac{(\deg_{\rm out}-\deg_{\rm in})^2}{(1-\kappa)\deg_{\rm
  out}}\, .
\end{align}
Using the fact that the degree outside $S$ is Poisson with mean
$\deg_{\rm out} = b$, and inside is Poisson with mean $\deg_{\rm in}
=\kappa a+(1-\kappa) b$, we also have
\begin{align}
\lambda = \frac{\kappa^2(a-b)^2}{(1-\kappa)b} \, .
\end{align}
We will therefore think in terms of the three  independent parameters:
$\kappa$ (the relative size of $|S|$); $b$ (the average degree in the
background); $\lambda$ (the signal-to-noise ratio).

We generically find two solutions of the cavity recursion, that possibly
coincide depending on the parameters values. This correspond to two
distinct phases of the statistical mechanics models, and also have a
useful algorithmic interpretation, which will be spelled out in detail
in Section \ref{sec:TreeInterpretation}.

Initializing the recursion with the  `exact solution' of the
reconstruction problem (`plus' initialization),  we  converges to a ferromagnetic fixed
point. This provides an upper bound on the performance of any
reconstruction algorithm.
Initializing the recursion with a completely oblivious initialization
(`free' initialization),
we converge to a paramagnetic fixed point. This also corresponds to 
the performance of the best possible local algorithm (see next section
for a formal definition).
A very similar qualitative picture is found in  other inference
problems on random graphs, one early example being the analysis of
sparse graph codes \cite{RiU08,MezardMontanari}.
An important simplification is that we do not expect replica-symmetry
breaking in these models \cite{NishimoriBook,MontanariSparse}. 

Depending on the model parameters, we encounter two types of 
behaviors as $\lambda$ increases. 
\begin{itemize}
\item For large $\kappa$ or small $b$,
the two fixed points mentioned above coincide for all $\lambda$ and 
no phase transition takes place. 
\item For small $\kappa$ and large $b$, two phase transitions take 
place:  a static phase transition at $\lambda_{\rm s}(\kappa,b)$ and a
dynamic phase transition at a larger value $\lambda_{\rm
  d}(\kappa,b)$. 
In addition, a spinodal point occurs at $\lambda_{\rm
  sp}(\kappa,b)<\lambda_{\rm s}(\kappa,b)<\lambda_{\rm d}(\kappa,b)$. 

For $\lambda<\lambda_{\rm sp}$ the two fixed point above coincide, and
yield bad reconstruction. For $\lambda>\lambda_{\rm d}$ they coincide
and yield good reconstruction. In the intermediate phase $\lambda_{\rm
sp}\le \lambda \le \lambda_{\rm d}$, the two
fixed points do not coincide. The relevant fixed point for
Bayes-optimal reconstruction corresponds to
the one of smaller free energy, and the transition between the two
takes place at $\lambda_{\rm s}$.
\end{itemize}
The reader might consult Fig. \ref{fig:PhaseDiagLargeDeg} for an
illustration. Also, a very similar phase diagram was obtained in
the related problem of sparse principal component analysis in \cite{deshpande2014information,lesieur2015phase}.

\subsection{Cavity equations and population dynamics}

Fixing $i$, let  $\prob(\,\cdot\,|i\in S)$ (respectively $\prob(\,\cdot\,|i\not\in S)$) be the law of 
$G$ subject to $S$ containing (respectively --not containing) vertex $i$.
Consider the random variable
\begin{align}
\xi_i(G) \equiv \log \frac{\prob(x_i=1|G)}{\prob(x_i=0|G)}\, .
\end{align}
The likelihood ratio test (maximizing $\Psucc$) amounts to
choosing\footnote{Another natural choice would be to minimize 
$\prob(T_i(G)\neq x_i)$. This is achieved by setting $T_i(G) =
\ind(\xi_i(G)\ge 0)$.}
\begin{align}
T^{\rm opt}_i(G) = \ind\Big(\xi_i(G)\ge \log \frac{\kappa}{1-\kappa}\Big)\,. 
\end{align}

As $N\to\infty$, the distribution of $\xi_i(G)$ under
$\prob(\,\cdot\,|i\in S)$ converges to the law of a certain random
variable $\xi_1$, and the distribution of $\xi_i(G)$ under
$\prob(\,\cdot\,|i\not\in S)$  converges instead to $\xi_0$.
The cavity method allows to write fixed point equations for these 
limit distributions. We omit details of the derivation since they are
straightforward given the model (\ref{eq:ApproxModel}), and since it
is
sufficient here to consider the replica-symmetric version of the method. 
General derivations can be found in \cite[Chapter 14]{MezardMontanari}.
A closely related  calculation is carried out in
\cite{decelle2011asymptotic},
which studies a more general random graph model, the so-called
stochastic block model.

The distribution of $\xi_1$, $\xi_0$ are fixed point of the following
recursion (the symbol $\ed$ means that the
distributions of quantities on the two sides are equal)
\begin{align}
\xi_0^{(t+1)} &\ed h + \sum_{i=1}^{L_{00}} f(\xi^{(t)}_{0,i}) +
\sum_{i=1}^{L_{01}} f(\xi^{(t)}_{1,i})\, ,\label{eq:GeneralCavity1}\\ 
\xi_1^{(t+1)} &\ed h + \sum_{i=1}^{L_{10}} f(\xi^{(t)}_{0,i}) +
\sum_{i=1}^{L_{11}} f(\xi^{(t)}_{1,i})\, . \label{eq:GeneralCavity2}\
\end{align}
Here $\xi^{(t)}_{0/1,i}$ are independent copies of
$\xi^{(t)}_{0/1}$. Further 
$L_{00}\sim \Poisson((1-\kappa)b)$, 
$L_{01}\sim \Poisson(\kappa b)$, 
$L_{10}\sim \Poisson((1-\kappa)b)$, 
$L_{11}\sim \Poisson(\kappa a)$,  are independent Poisson random variables, independent
of the $\{\xi^{(t)}_{0,i}\}$. 
Finally, 
\begin{align}
h = \log \gamma = -\kappa(a-b)-
\log\Big(\frac{1-\kappa}{\kappa}\Big)\, ,
\end{align}
and the function $f:\reals\to\reals$ is given by
\begin{align}
f(\xi) \equiv \log\Big(\frac{1+\rho \, e^{\xi}}{1+ e^{\xi}}\Big)\, .
\end{align}
(Recall that $\rho = a/b$, cf. Eq.~(\ref{eq:RhoDef}).)

The cavity method predicts that the asymptotic distribution of
$\xi_i(G)$ (conditional to $i\in S$ or $i\not\in S$) is a fixed point
of Eqs.~(\ref{eq:GeneralCavity1}), (\ref{eq:GeneralCavity2}).
In order to find the fixed points, we iterate these distributional
equations with two types of initial conditions (that correspond,
respectively, to the poor reconstruction and good reconstruction phases)
\begin{align}
{\rm  free}\;\;:&\;\;\; \begin{cases}
\xi^{(0),\fr}_{0} = \log(\kappa/(1-\kappa))\, ,\\
\xi^{(0),\fr}_{1} = \log(\kappa/(1-\kappa))\, ,
\end{cases}\label{eq:FreeBc}\\
{\rm  plus}\;\;:&\;\;\; \begin{cases}
\xi^{(0),\pl}_{0} = -\infty\, ,\\
\xi^{(0),\pl}_{1} = +\infty\, .
\end{cases}\label{eq:PlusBc}
\end{align}
We refer to Section
\ref{sec:TreeInterpretation} for the interpretation and monotonicity
properties of these conditions: in particular it can be proved that
$\xi^{(t)}_{0/1}$ converge in distribution if initialized in this
manner. 
We implemented Eqs.~(\ref{eq:GeneralCavity1}),
(\ref{eq:GeneralCavity2})
numerically using the `population dynamics' method\footnote{As a
  technical parenthesis, we found it useful to impose the constraint
  $\E(x_i)=\kappa$ in the sampled density evolution. This was done
  using the method of \cite{di2004weight}.} of \cite{mezard2001bethe} (also
known as `sampled density evolution' \cite{RiU08,MezardMontanari}).

\begin{figure}[t!]
 \includegraphics[scale=0.4]{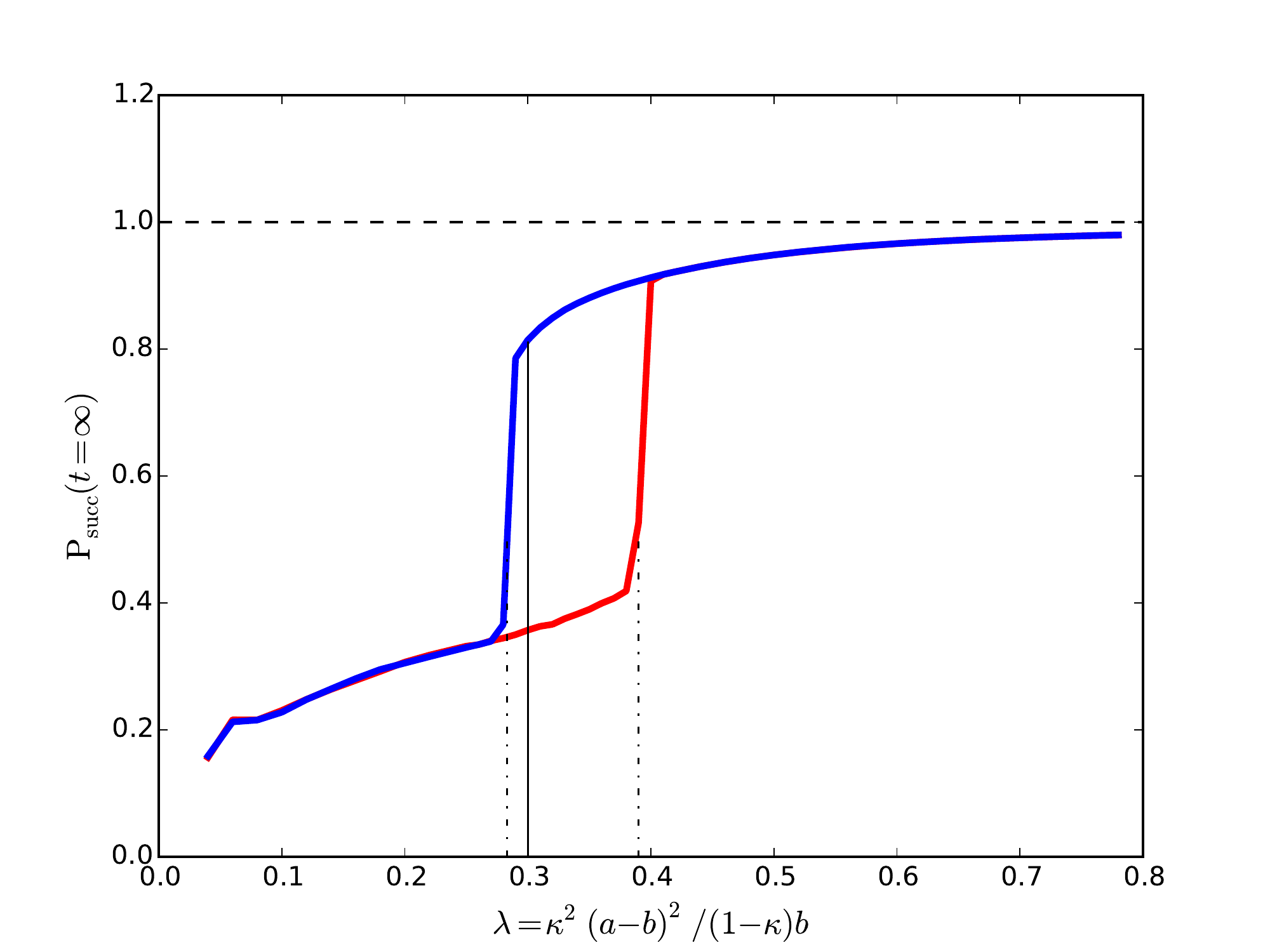}
\hspace{-0.2cm}  \includegraphics[scale=0.4]{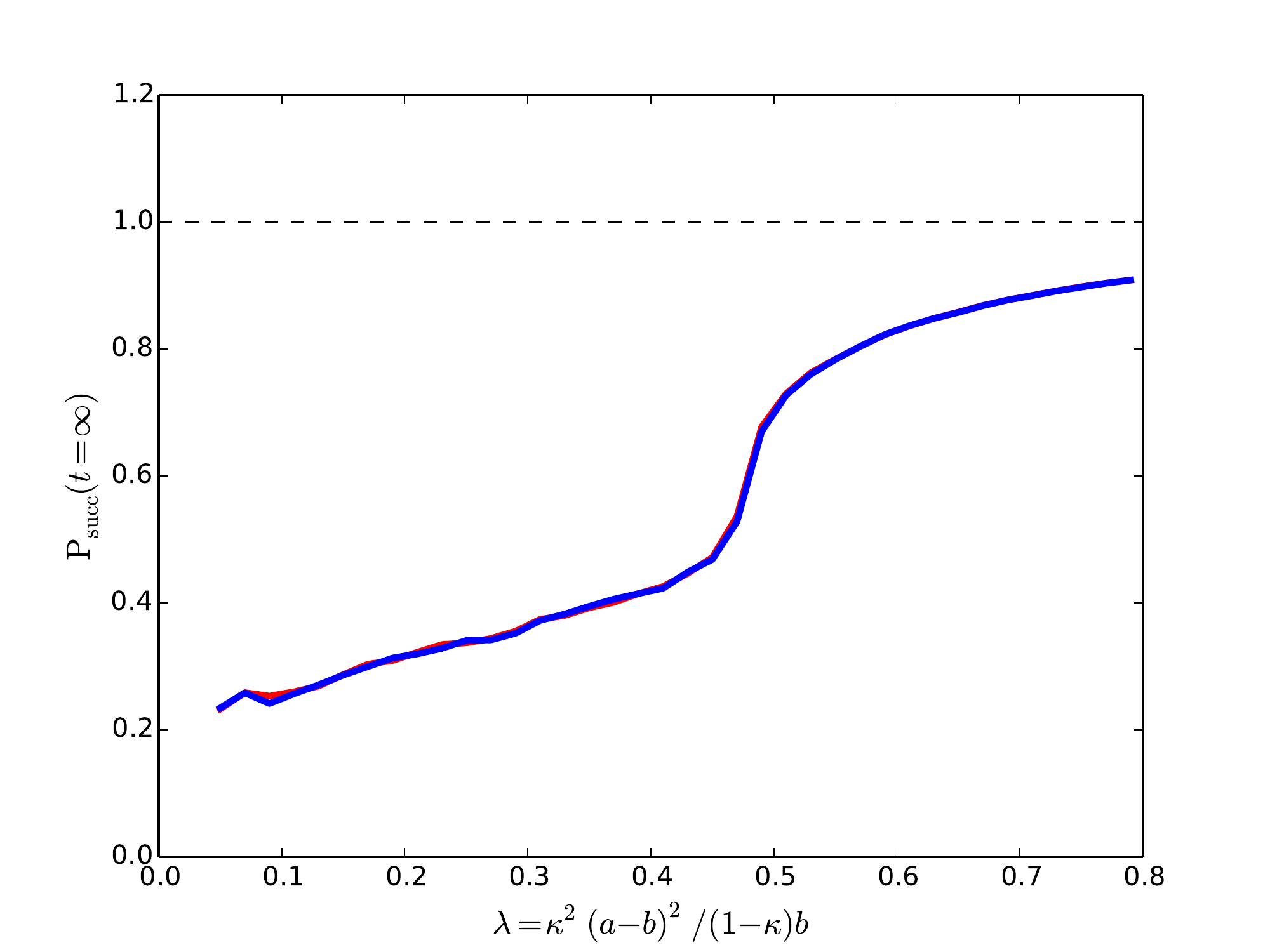}
\put(-385,23){{\tiny $\lambda_{\rm sp}$}}
\put(-365,23){{\tiny $\lambda_{\rm s}$}}
\put(-345,23){{\tiny $\lambda_{\rm d}$}}
\caption{The success probability in the two different phases,
for $\kappa = 0.005$ (left), $0.020$ (right) and $b=100$
(corresponding to average degree outside the set $S$, $\deg_{\rm out}=100$).
Red curves correspond $\Psucc(\frl)$ (i.e. free boundary/initial conditions), and
provide to the optimal performance of local algorithms. Blue curves
yield  $\Psucc(\pll)$ (i.e. plus boundary/initial conditions) and yield
an upper bound on the performance of any algorithm.
The continuous black line  at $\lambda_{\rm s}\approx 0.3$ coincides with
the phase transition of Bayes-optimal estimation.
These curves were computed by averaging over $10$ runs
of the population dynamics algorithm with
$M=10^4$ samples and  $300$ iterations.}
  \label{fig:SparsePsucc}
\end{figure}
In Figure \ref{fig:SparsePsucc}, we plot the predicted behavior of $\Psucc$ for
$b=100$  and two different values of the clique size: $\kappa \in
\{0.005, 0.020\}$. The success probability is predicted to
be (for $N\to\infty$)
\begin{align}
\Psucc = \prob(\xi^{(\infty)}_1\ge 0) + \prob(\xi_0^{(\infty)}<0)-1\, .
\end{align}
We denote by $ \Psucc(\frl)$ and $\Psucc(\pll)$ the predictions obtained
with the two initializations above.

As anticipated two behaviors can be observed. 
For $\kappa$ sufficiently large, the curves $\Psucc(\frl)$ and
$\Psucc(\pll)$ coincide for all $\lambda$. When this happens, this is
also the success probability of the optimal likelihood ratio test
$T^{\rm opt}$, and the latter can be effectively approximated using a
local algorithm (e.g. belief propagation), see Section
\ref{sec:TreeInterpretation}.
For $\kappa$ small the two curves remain distinct in an intermediate
interval of values: $\lambda\in (\lambda_{\rm sp},\lambda_{\rm d})$.

In this regime, the asymptotic behavior of the Bayes-optimal test 
is captured by the fixed point that yields the lowest free energy.
It is convenient to define the rescaled free energy density as follows
(assuming that the limit exists)
\begin{align}
\free \equiv \frac{\kappa^2}{2}\, \Big(a\log \frac{a}{b}-2a+2b\Big)-\log(1-\kappa)-
\lim_{n\to\infty}\frac{1}{n}\E\log Z(G)\, .
\end{align}
The reason for this choice of the additive constants is
that the resulting free energy is also equal to the asymptotic mutual
information between the hidden set $S$ and the observed graph $G$
\begin{align}
\free = \lim_{n\to\infty} \frac{1}{N}\ \Info(G;S)\, .
\end{align}
This quantity has therefore an immediate interpretation and several
useful properties. 

The replica symmetric cavity method (equivalently, Bethe-Peierls
approximation) predicts
\begin{align}
\free  = \min_{\P_0,\P_1} \Free(\P_0,\P_1) \, ,
\end{align}
where the supremum is over all probability distributions $\P_0,\P_1$
over the real line satisfying the following symmetry property (see
Section \ref{sec:TreeInterpretation} for further clarification on this
property):
\begin{align}
\frac{\de\P_1}{\de \P_0}(\xi) = \frac{1-\kappa}{\kappa} \, e^{\xi}\, .  \label{eq:Symmetry}
\end{align} 
The functional $\Free$ is defined as follows
\begin{align}
\Free &= \Pe - \Pv+\Free_0\, ,\label{eq:FreeEnergy}\\
\Pe & = \frac{1}{2}\big(\kappa^2
a+(1-\kappa^2)b\big)\,
\E\log\Big\{1+\frac{(\rho-1)\, e^{\xi_{x_1,1}+\xi'_{x_2,2}} }{(1+e^{\xi_{x_1,1}})(1+e^{\xi_{x_2,2}}))}\Big\}\,,\\
\Pv & = \E\log \Big\{1-\kappa+\kappa\, e^{-\kappa(a-b)}
\prod_{i=1}^{L_0} \Big(\frac{1+\rho\, e^{\xi_{0,i}}}{1+e^{\xi_{0,i}}}\Big)\prod_{j=1}^{L_1}
\Big(\frac{1+\rho\, e^{\xi_{1,j}}}{1+e^{\xi_{1,j}}}\Big)
\Big\}\, ,\\
\Free_0 &= \frac{\kappa^2}{2}\, \Big(a\log
\frac{a}{b}-2a+2b\Big)\, .
\end{align}
Here expectation is taken with respect to the following independent
random variables:
\begin{itemize}
\item $\{\xi_{0,i}\}$  that are
i.i.d. random variables with distribution $\P_0$; 
\item $\{\xi_{1,i}\}$  that are
i.i.d. random variables with distribution $\P_1$; 
\item  $(x_1,x_2)\in\{0,1\}^2$ with joint distribution $p_{1,1} =
  \kappa^2a/z$, $p_{0,1} = p_{1,0} = \kappa(1-\kappa)b/z$, $p_{1,1} =
  (1-\kappa)^2b/z$, where $z = \kappa^2 a+ (1-\kappa^2)b$. 
\item $(L_0,L_1)$ with the following mixture distribution.
With probability $\kappa$: $L_0\sim \Poisson((1-\kappa)b)$, $L_1\sim
\Poisson(\kappa a)$. With probability $(1-\kappa)$: $L_0\sim \Poisson((1-\kappa)b)$, $L_1\sim
\Poisson(\kappa b)$.
\end{itemize}
\begin{figure}[t!]
\hspace{4cm} \includegraphics[scale=0.4]{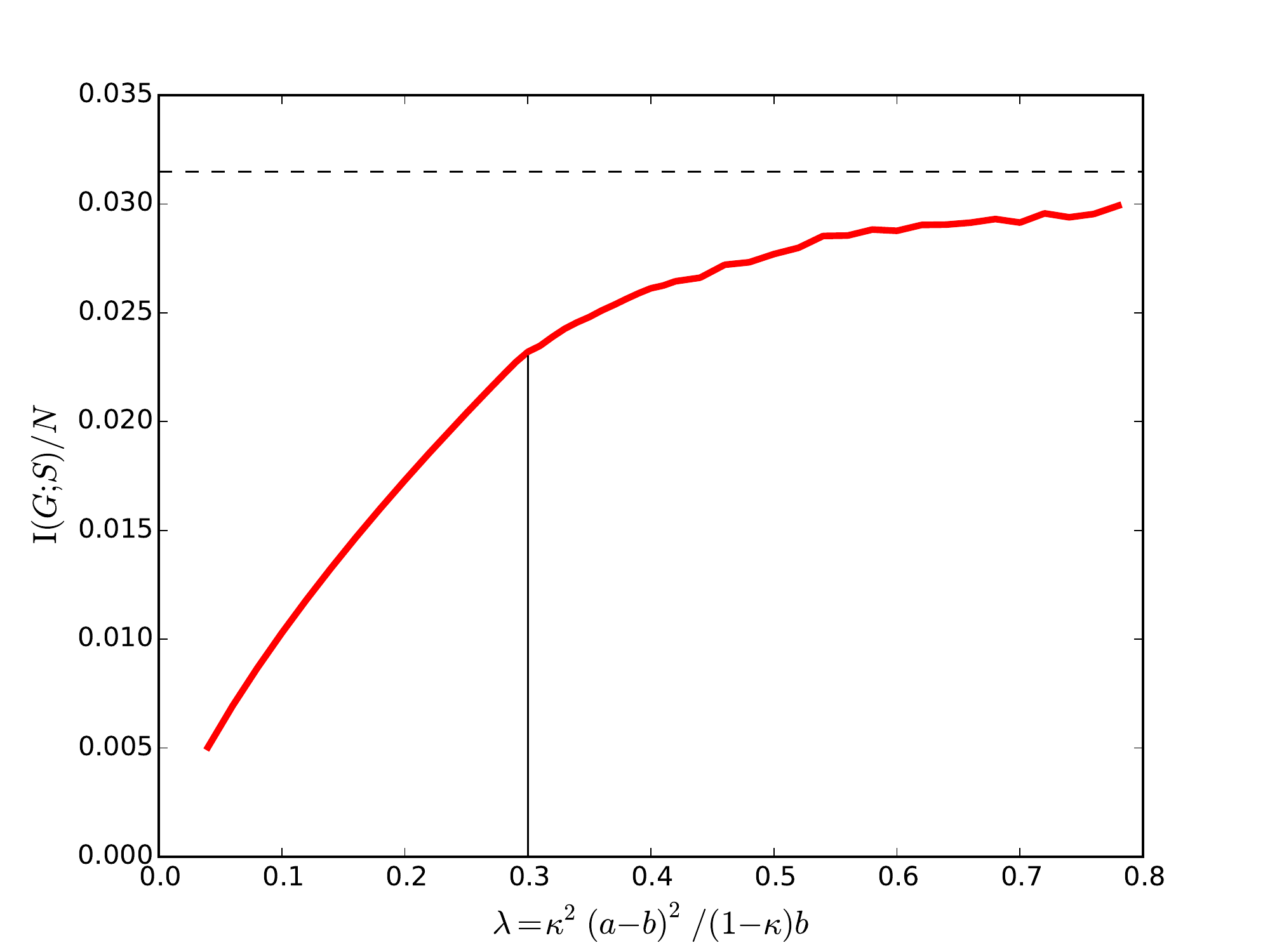}
\put(-145,23){{\tiny $\lambda_{\rm s}$}}
 \caption{The free energy density (equivalently, the mutual information
   per vertex),
for $\kappa = 0.005$. The horizontal line corresponds to the 
maximal mutual information $H(\kappa)\approx 0.03148$.,
The vertical line at $\lambda_{\rm s}\approx 0.3$ corresponds to the phase transition of the
Bayes-optimal estimator.
This curve was computed by averaging over $10$ runs
of the population dynamics algorithm with
$M=10^4$ samples and  $300$ iterations.}
  \label{fig:SparseFree}
\end{figure}
Let $\P_{0/1}^{\pl}$ and $\P_{0/1}^{\fr}$ the distributions of the
fixed points obtained with plus and free initial conditions.
In Figure \ref{fig:SparseFree} we plot the minimum of the corresponding Bethe free energies 
$\Free(\pll) =  \Free(\P_0^{\pl},\P_1^{\pl})$ and $\Free(\frl) =  \Free(\P_0^{\fr},\P_1^{\fr})$
for $b=100$, $\kappa = 0.005$ (as obtained by the population dynamics
algorithm). 
This is the cavity prediction
for  the free energy density $\free$. The value of $\lambda$ for which 
$\Free(\pll) =   \Free(\frl)$ corresponds to the phase transition point
$\lambda_{\rm s}$ between paramagnetic and ferromagnetic phases. 
From the reconstruction point of view, this is the phase transition
for Bayes-optimal estimation: 
\begin{align}
\lim_{N\to\infty}\Psucc^{(N)}(T^{\rm opt}) = 
\begin{cases}
\Psucc(\frl) & \mbox{ for $\lambda<\lambda_{\rm s}$}\, ,\\
\Psucc(\pll) & \mbox{ for $\lambda>\lambda_{\rm s}$}\, .
\end{cases}
\end{align}

Notice from Figure \ref{fig:SparseFree} that as expected  $\psi = \lim_{N\to\infty}\Info(G;S)/N$ is
monotone increasing in the signal-to-noise ratio $\lambda$, with
$\psi\to 0$ as $\lambda\to 0$, and $\psi\to H(\kappa)$ as
$\lambda\to\infty$
(here $H(\kappa) = -\kappa\log \kappa-(1-\kappa)\log (1-\kappa)$ is
the entropy of
a Bernoulli random variable with mean $\kappa$). Also, the curve
Fig. \ref{fig:SparseFree} presents some `wiggles' at large $\kappa$
that are due to the limited numerical accuracy of the population
dynamics algorithm.

\subsection{Large-degree asymptotics}

In the previous section we solved numerically
the distributional equations (\ref{eq:GeneralCavity1}),
(\ref{eq:GeneralCavity2}).
This approach is somewhat laborious and its accuracy is limited. Asymptotic expansions provide 
complementary analytical insights into the solution of these equations.
\begin{figure}[t!]
\phantom{a}\hspace{-0.75cm}\includegraphics[scale=0.3]{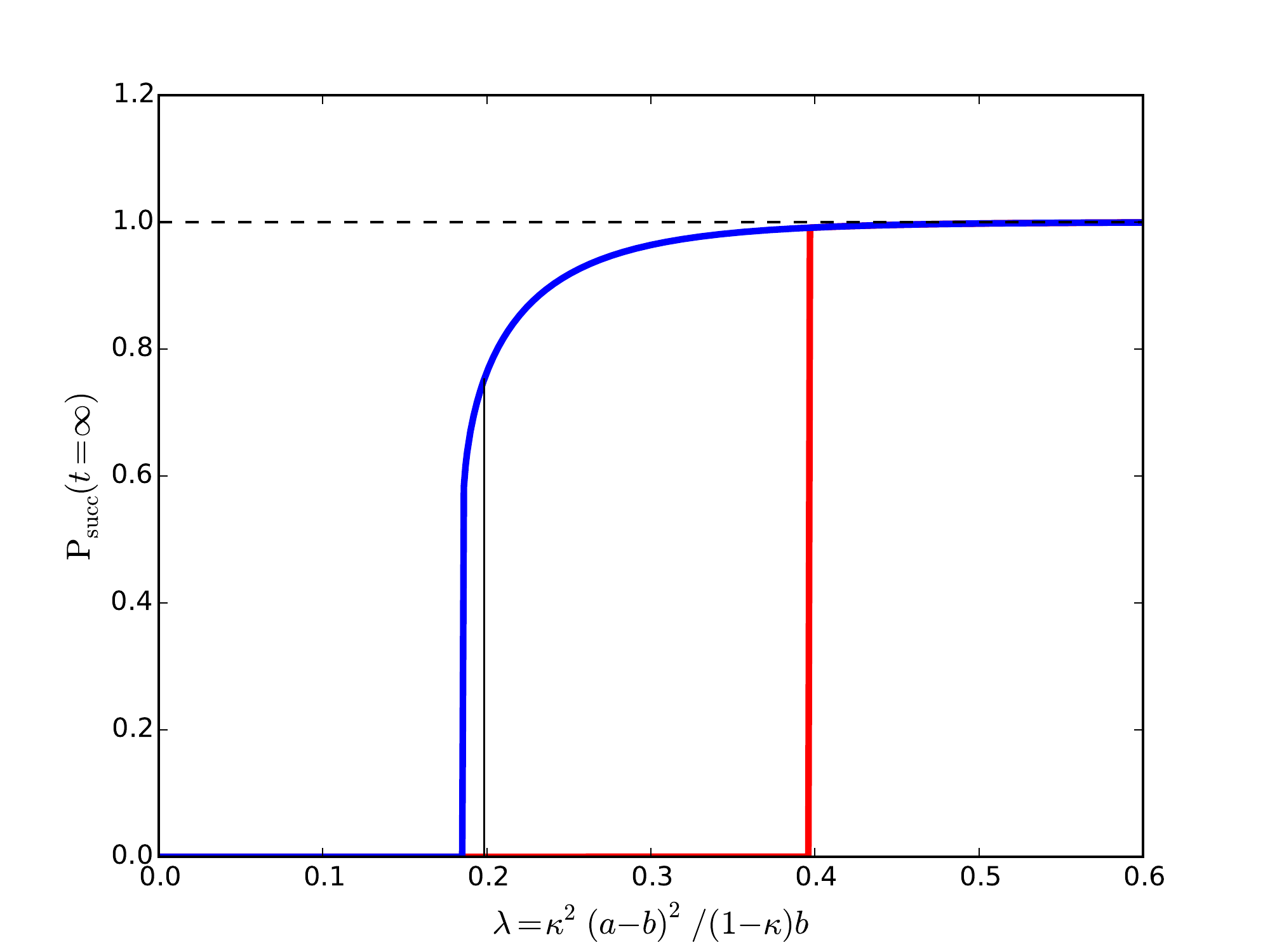}
\hspace{-0.5cm}\includegraphics[scale=0.3]{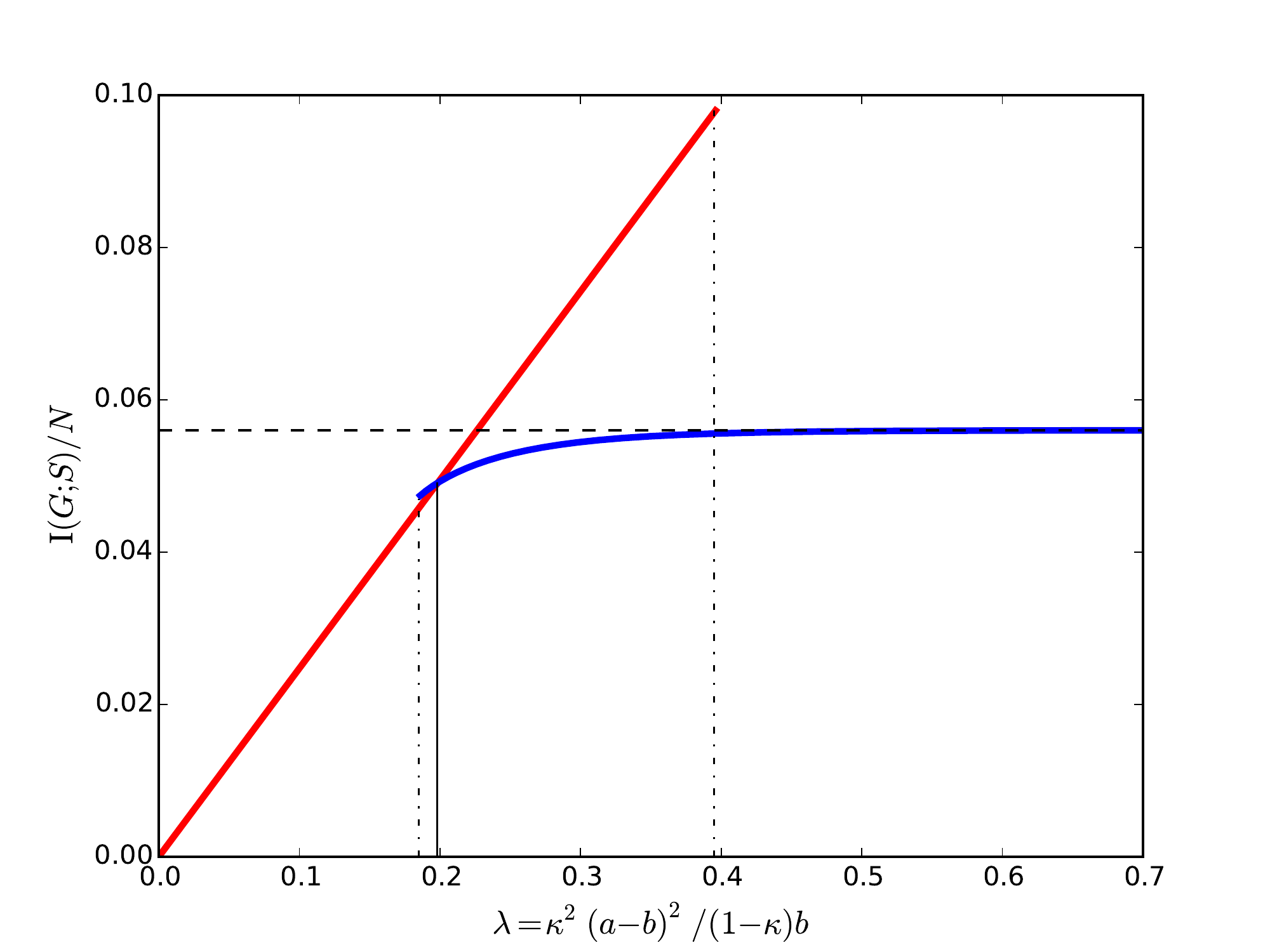}
\hspace{-0.5cm}\includegraphics[scale=0.3]{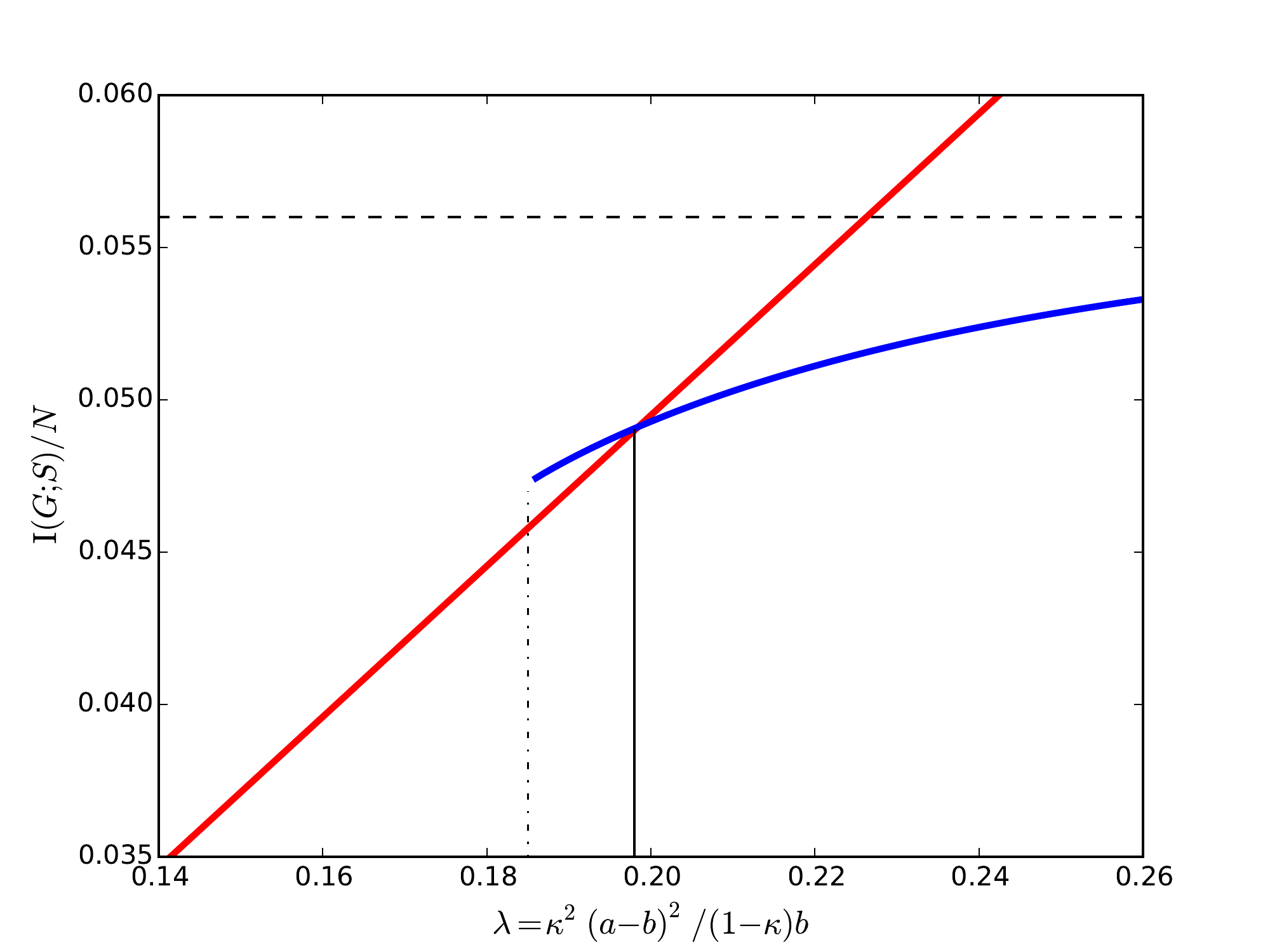}
\put(-450,17){{\tiny $\lambda_{\rm sp}$}}
\put(-423,17){{\tiny $\lambda_{\rm s}$}}
\put(-380,17){{\tiny $\lambda_{\rm d}$}}
\put(-292,17){{\tiny $\lambda_{\rm sp}$}}
\put(-270,17){{\tiny $\lambda_{\rm s}$}}
\put(-233,17){{\tiny $\lambda_{\rm d}$}}
\put(-120,17){{\tiny $\lambda_{\rm sp}$}}
\put(-80,17){{\tiny $\lambda_{\rm s}$}}
\caption{Limit $a,b\to\infty$ with $\lambda$ and $\kappa$ fixed. Here
  $\kappa=0.01$. Left frame: Success probability for free boundary
  condition (equivalently, local algorithms, red curve), and plus
  boundary condition (equivalently, general upper bound, blue
  curve). Center frame: free energy (equivalently, mutual information
  per vertex) with same boundary conditions. Right frame: zoom of the
  free energy curves.}\label{fig:Imu}
\end{figure}
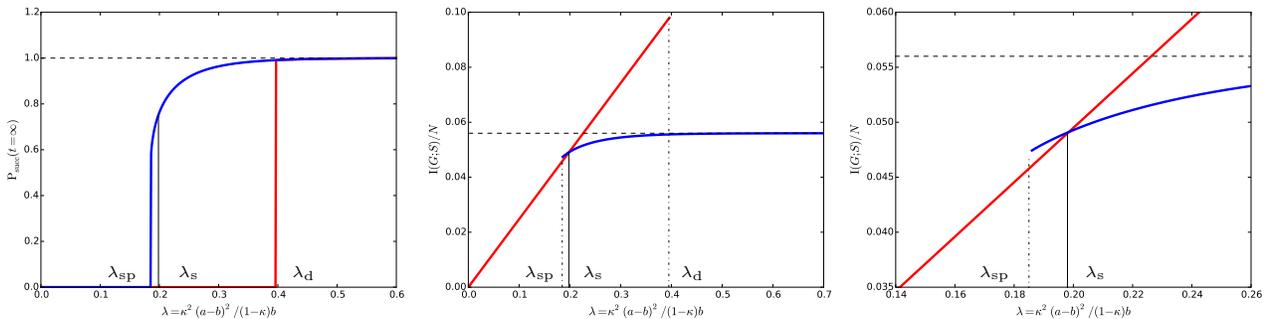

Here we consider $a,b\to\infty$ with $\kappa$ fixed, and $(a-b)/b^2$ converging to
a limit. In particular, the signal-to-noise ratio $\lambda$ is also a
constant.
Let us emphasize once more that these limits are taken \emph{after}
$N\to\infty$
and hence the graph is still sparse.

\begin{figure}[t!]
\begin{center}
\includegraphics[scale=0.45]{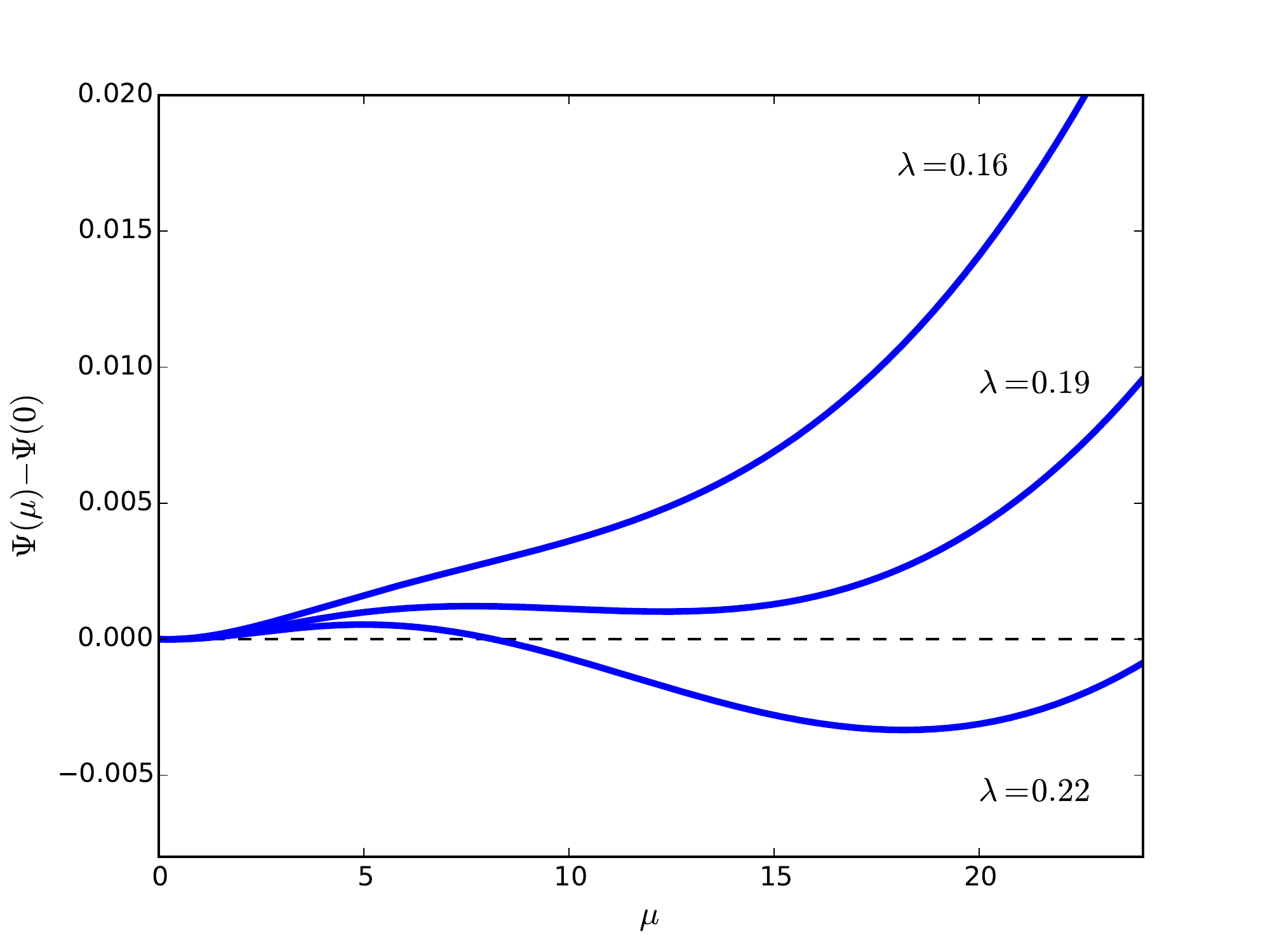}
\end{center}
\caption{Limit $a,b\to\infty$ with $\lambda$ and $\kappa$ fixed. Here
we plot the (shifted)  free energy function $\Psi(\mu)-\Psi(0)$ for 
  $\kappa=0.01$ and $\lambda\in\{0.16,0.19,0.22\}$. Comparing with
  Figure \ref{fig:Imu} we see that $0.16<\lambda_{\rm
    sp}(\kappa)<\lambda_{\rm s}(\kappa)$, $\lambda_{\rm
    sp}(\kappa)<0.19<\lambda_{\rm s}(\kappa)$, $\lambda_{\rm
    sp}(\kappa)<\lambda_{\rm s}(\kappa)<0.22<\lambda_{\rm d}(\kappa)$.}\label{fig:PsiMu}
\end{figure}
In this limit, the fixed points of Eqs.~(\ref{eq:GeneralCavity1}),
(\ref{eq:GeneralCavity2}) take the form
\begin{align}
\xi_0 & \sim
\normal\Big(-\log\Big(\frac{1-\kappa}{\kappa}\Big)-\frac{1}{2}\,\mu,\;
\mu\Big)\, ,\label{NormalApprox1}\\
\xi_1 & \sim
\normal\Big(-\log\Big(\frac{1-\kappa}{\kappa}\Big)+\frac{1}{2}\,\mu,\;
\mu\Big)\, . \label{NormalApprox2}
\end{align}
Further $\mu$ satisfies the fixed point equation
\begin{align}
\mu = \lambda \, \F(\mu;\kappa) \label{eq:CavityLargeDegree}
\end{align}
where the function $\F(\,\cdot\,;\,\cdot\,)$ is defined by
\begin{align}
\F(\mu;\kappa) \equiv \E\Big\{\frac{1-\kappa}{\kappa + (1-\kappa)
  e^{-(\mu/2)+\sqrt{\mu} \, Z } }\Big\}\, ,\label{eq:CavityLargeDegreeLast}
\end{align}
with expectation being taken with respect to $Z\sim\normal(0,1)$.
In other words, the distributional equations (\ref{eq:GeneralCavity1}),
(\ref{eq:GeneralCavity2}) reduced to a single nonlinear equation for
the scalar $\mu$. Large $\mu$ correspond to accurate recovery. 

More formally, we expect the distributional solutions of Eqs.~(\ref{eq:GeneralCavity1}),
(\ref{eq:GeneralCavity2}) to converge to solutions of
Eqs.~(\ref{NormalApprox1}) to (\ref{eq:CavityLargeDegreeLast}).
We do not provide a `physicists' derivation of this statement since
this follows heuristically \footnote{Of course  Lemma \ref{lemma:GaussianApprox} does not prove rigorously that the fixed points of
Eqs.~(\ref{eq:GeneralCavity1}),
(\ref{eq:GeneralCavity2}) converge to fixed points of  Eqs.~(\ref{NormalApprox1}) to
(\ref{eq:CavityLargeDegreeLast}). A complete
proof would require controlling the convergence rate to fixed
points. However in heuristic statistical physics derivation this is
typically not  done. Also, the proof of Lemma
\ref{lemma:GaussianApprox} follows the same strategy that would be
employed in a heuristic derivation.} from  Lemma \ref{lemma:GaussianApprox}. The
latter establishes that, iterating  the cavity equations Eqs.~(\ref{eq:GeneralCavity1}),
(\ref{eq:GeneralCavity2}) any fixed number of times $t$ is equivalent
(in the large-degree limit) to iterating Eqs.~(\ref{NormalApprox1}) to
(\ref{eq:CavityLargeDegreeLast}).

\begin{figure}[t!]
\centering
  \includegraphics[scale=0.5]{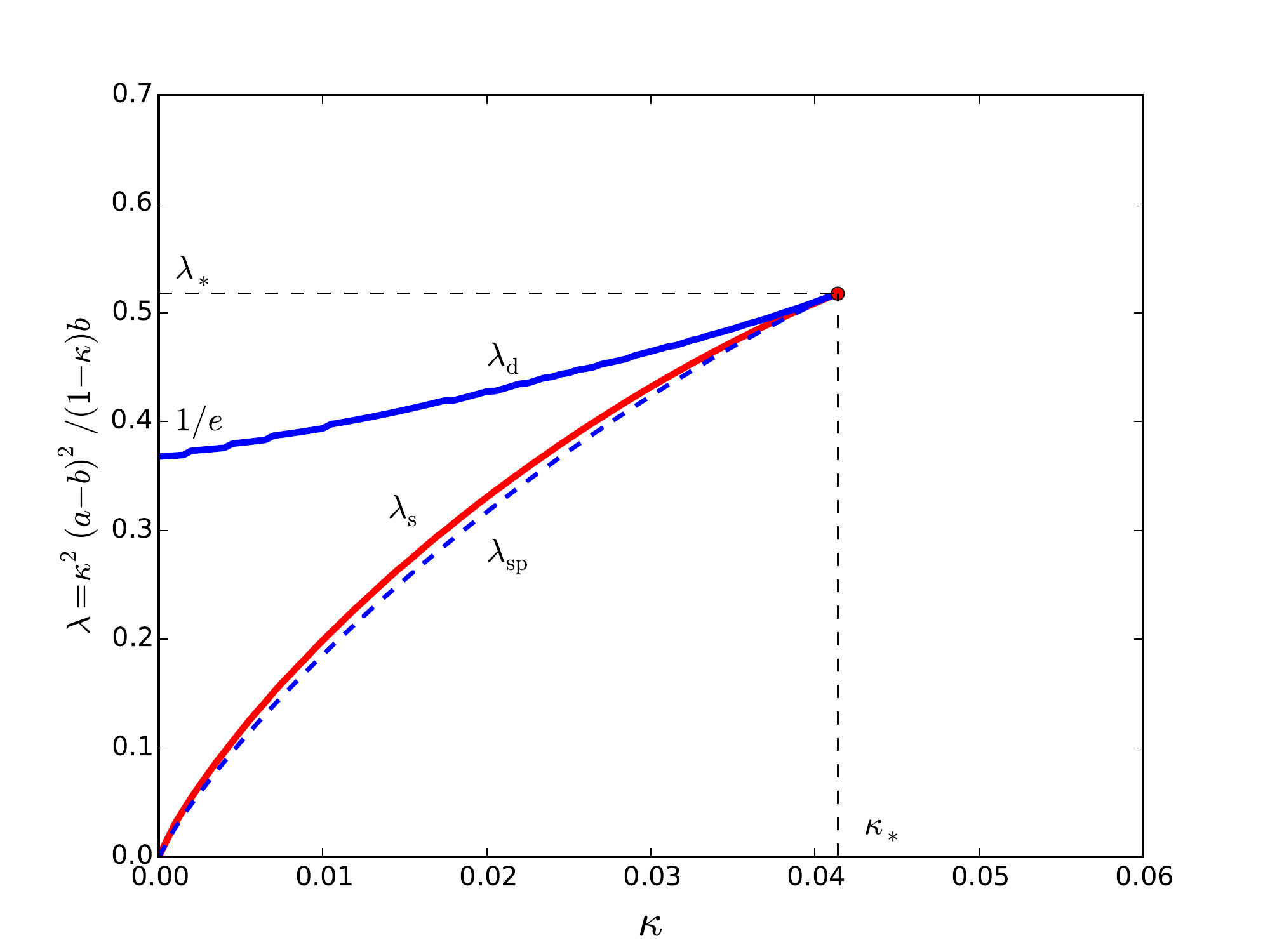}
 \caption{Phase diagram of the hidden subgraph problem in the large
   degree limit $a,b\to\infty$, with $\kappa = \E|S|/N$ (relative
   size of the hidden set) and $\lambda =
   \kappa^2(a-b)^2/((1-\kappa)b)$ (signal-to-noise ratio) fixed. The three curves are, from
   top to bottom $\lambda_{\rm d}(\kappa)$, $\lambda_{\rm s}(\kappa)$
   and $\lambda_{\rm sp}(\kappa)$.}
  \label{fig:PhaseDiagLargeDeg}
\end{figure}
The free energy (\ref{eq:FreeEnergy}) becomes a function of $\mu$ (we
still denote it by $\Free$ with a slight abuse of notation):
\begin{align}
\Free(\mu) &= \frac{1}{4}\lambda(1-\kappa)+\frac{\kappa^2}{4\lambda(1-\kappa)}\mu^2-\E\log\Big\{1-\kappa+\kappa\,\exp\Big(\sqrt{\mu}Z-\frac{1}{2}\mu+\mu
X\Big)\Big\}\, ,
\end{align}
where expectation is with respect to independent random variables
$X\sim\Ber(\kappa)$
and $Z\sim\normal(0,1)$.
Its local minima are solutions of Eq.~(\ref{eq:CavityLargeDegree}).

Equation (\ref{eq:CavityLargeDegree}) can be easily solved
numerically, yielding the phase diagram in Figure
\ref{fig:PhaseDiagLargeDeg}.
As before, we obtain phase transitions $\lambda_{\rm
  sp}(\kappa)<\lambda_{\rm s}(\kappa)<\lambda_{\rm d}(\kappa)$
as long as $\kappa$ is below a critical point $\kappa<\kappa_*$.
The critical point location is
\begin{align}
\kappa_* \approx 0.04139\, ,\;\;\;\;\;\;\;\;\;
\lambda_*\approx 0.5176\, .
\end{align}
The free energy $\Free(\mu)$ has two local minima $\mu^{\pl}>\mu^{\fr}$
for
$\kappa<\kappa_*$,
$\lambda\in (\lambda_{\rm sp}(\kappa),\lambda_{\rm d}(\kappa))$, and
one local minimum otherwise.
The local minimum $\mu^{\pl}$ is the global minimum for
$\lambda>\lambda_{\rm s}(\kappa)$, while $\mu^{\fr}$ is the global minimum for
$\lambda<\lambda_{\rm s}(\kappa)$.
We refer to Figures \ref{fig:Imu} and \ref{fig:PsiMu} for illustration.

Of particular interest is the case of small hidden subsets, i.e. the
limit $\kappa\to 0$ (note that $|S|$ is still linear  in $N$).
For small $\kappa$ we have $\lim_{\kappa\to 0}\F(\mu;\kappa) = \F(\mu;0) = e^{\mu}$.
Hence the solutions (\ref{eq:CavityLargeDegree}) that stay bounded
converges to the solution of
\begin{align}
\mu = \lambda\, e^{\mu}\, .
\end{align}
This equation has two solutions for $\lambda<1/e$ and no solution for
$\lambda>1/e$.
This implies that
\begin{align}
\lim_{\kappa\to 0}\lambda_{\rm d}(\kappa) = \frac{1}{e}\, ,
\end{align}
which is the result announced in Eq.~(\ref{eq:BasicCondition}).
It is also easy to see that $\lambda_{\rm s}(\kappa),\lambda_{\rm
  sp}(\kappa)\to 0$ as $\kappa\to 0$.

\subsection{Algorithmic interpretation}
\label{sec:TreeInterpretation}

The distributional equations (\ref{eq:GeneralCavity1}) and
(\ref{eq:GeneralCavity2}) 
define a sequence  of probability distributions indexed by
$t\in\{0,1,2,\dots\}$. More precisely, for every $t$  the recursion
defines the probability distributions $\P_0^{(t)}$ (the distribution of
$\xi_0^{(t)}$) and $\P_1^{(t)}$ (the distribution of
$\xi_1^{(t)}$).
When specialized to the free/plus initial conditions
(cf. Eqs. ~(\ref{eq:FreeBc}), ~(\ref{eq:FreeBc})), these probability
distributions have a simple and useful interpretation that we will now
explain\footnote{The
  discussion follows very closely what happens in other inference
  problem, for instance in the analysis of sparse graph codes \cite{MezardMontanari,RiU08}.}.

Define $\Ball_t(G,i)$ to be the ball of radius $t$ centered at $i\in
V$, in graph $G$. Namely, this is the subset of vertices of $G$ whose
distance from $i$ is at most $t$. By a slight abuse of notation,
this will also denote the subgraph induced in $G$ by those vertices.
The following remarks are straightforward.

\vspace{0.2cm}

\noindent{\bf Free boundary condition.} Consider the optimal test
$T_i(G)$ among those that only use local information. In other words, 
$T_i(G)$ is the optimal test that is a function of
$\Ball_t(G,i)$. This is again a likelihood ratio test. Concretely, we
can define the log-likelihood ratio
\begin{align}
\xi_i(G;t) \equiv \log
\frac{\prob(x_i=1|\Ball_t(G,i))}{\prob(x_i=0|\Ball_t(G,i))}\, .
\end{align}
Then the optimal test takes the form $T_{i}(G) = \ind(\xi_i(G)\ge \log[\kappa/(1-\kappa)])$ 
(if we are interested in maximizing $\Psucc^{(N)}(T)$) or  $T_{i}(G) =
\ind(\xi_i(G)\ge 0)$ 
(if we are interested in minimizing the expected number of incorrectly
assigned vertices).

Fixing the depth parameter $t$, the distribution of $\xi_i(G;t)$
converges (as $N\to \infty$) to $\P_0^{(t),\fr}$ for $i\in S$,
and to  $\P_1^{(t),\fr}$ for $i\not\in S$. Mathematically, for any
fixed $i$ 
\begin{align}
\xi_i(G;t) &\stackrel{{\rm d}}{\Rightarrow} \xi_1^{(t),\fr}\;\;\;\;\;\; \mbox{
  under }\prob(\,\cdot\, |i\in S)\, ,\\
\xi_i(G;t) &\stackrel{{\rm d}}{\Rightarrow} \xi_0^{(t),\fr}\;\;\;\;\;\; \mbox{
  under }\prob(\,\cdot\, |i\not\in S)\, .
\end{align}
In particular, for any fixed $t$, the success probability 
\begin{align}
\Psucc(t;\frl) = \P_0^{(t),\fr}\Big(\xi<\log[\kappa/(1-\kappa)]\Big)
+\P_1^{(t),\fr}\Big(\xi\ge \log[\kappa/(1-\kappa)]\Big)-1\, \label{eq:PsuccFree}
\end{align}
is the maximum asymptotic success probability achieved by any test that is
$t$-local (in the sense of being a function of depth-$t$
neighborhoods). It follows immediately from the definition that
$\Psucc(t;\frl)$ is monotone increasing in $t$.
Its $t\to\infty$ limit $\Psucc(\frl)$ is the maximum success
probability  achieved by any local algorithm.
This  quantity was computed through
population dynamics in the previous section, see Figure \ref{fig:SparsePsucc}.

\vspace{0.2cm}

\noindent{\bf Plus boundary condition.} Let $\bBall_t(G,i)$ be the
complement of $\Ball_t(G,i)$, i.e. the set of vertices of $G$
that have distance at least $t$ from $i$. Then $\xi^{(t),\pl}_{0/1}$
has the interpretation of being the log-likelihood ratio, when
information is revealed about the labels of vertices in
$\bBall_{t-1}(G,i)$.
Namely, if we define
\begin{align}
\xi'_i(G;t) \equiv \log
\frac{\prob(x_i=1|\Ball_t(G,i),x_{\bBall_t(G,i)}
  )}{\prob(x_i=0|\Ball_t(G,i), x_{\bBall_t(G,i)} )}\, ,
\end{align}
then we have
\begin{align}
\xi'_i(G;t) &\stackrel{{\rm d}}{\Rightarrow} \xi_1^{(t),\pl}\;\;\;\;\;\; \mbox{
  under }\prob(\,\cdot\, |i\in S)\, ,\\
\xi'_i(G;t) &\stackrel{{\rm d}}{\Rightarrow} \xi_0^{(t),\pl}\;\;\;\;\;\; \mbox{
  under }\prob(\,\cdot\, |i\not\in S)\, .
\end{align}
In particular, $\Psucc(\pll)$  is an upper bound on the performance of
any estimator. In the previous section we computed this quantity
numerically through population dynamics.

\vspace{0.2cm}

Let us finally comment on the relation (\ref{eq:Symmetry}) between
$\P_0$ and $\P_1$.
This is an elementary consequence of Bayes formula:
 consequences of this relation have been useful  in statistical
 physics under the name of `Nishimori property' \cite{NishimoriBook}.
It is also known in coding theory as `symmetry condition'
\cite{RiU08}.
Consider the general setting of two random variables $X,Y$, with $X\in
\{0,1\}$, $\prob(X=1)=\kappa$, and let $\xi(Y)=
\log[\prob(X=1|Y)/\prob(X=0|Y)]$.
Then for any interval $A$ (with non-zero probability), applying Bayes formula,
\begin{align}
\prob(\xi(Y)\in A|X=1) &= 
\frac{\prob(\xi(Y)\in A;X=1)}{\prob(X=1)} \\
&=\frac{1}{\kappa} \, \E\big\{\ind(\xi(Y)\in A)\prob(X=1|Y)\big\}=
\frac{1}{\kappa} \, \E\big\{\ind(\xi(Y)\in A)\,\ind(X=0) e^{\xi(Y)} \big\}\\
&=\frac{1-\kappa}{\kappa} \,\E\big\{\ind(\xi(Y)\in A)\,  e^{\xi(Y)}
|X=0\big\}\, ,
\end{align}
which is the claimed property.

%
%
\section{Rigorous results}
\label{sec:Rigorous}

In the previous section we relied on the non-rigorous cavity method
from spin glass theory to derive the phase diagram. Most notably we 
used numerical methods, and  formal large-degree asymptotics
to study the distributional equations (\ref{eq:GeneralCavity1}),
(\ref{eq:GeneralCavity2}).
Here we will establish rigorously some key implications of the phase
diagram, namely:
\begin{itemize}
\item By exhaustive search over all subsets of $k$ vertices in $G$,
we can estimate $S$ accurately for any $\lambda>0$ and $\kappa$ small.
\item Local algorithms succeed in reconstructing accurately $S$ 
 if $\lambda>1/e$, and fail for $\lambda<1/e$ (assuming large degrees
 and  $\kappa$ small).
\end{itemize}
\subsection{Exhaustive search}

Given a set of vertices $R\subseteq[N]$, we let $E(R)$ denote the
number of edges with both endpoints in $R$. Exhaustive search
maximizes this quantity among all the sets that have the `right size.'
Namely, it outputs
\begin{align}
\hS = \arg\max_{R\subseteq [N]}\Big\{\,  E(R)\, : |R| = \lfloor\kappa n \rfloor\Big\}\, .
\end{align}
(If multiple maximizers exist, one of them is selected arbitrarily.)
We can also define a test function $T_i(G)$ by letting
$T^{\ex}_{i}(G) =1$ for $i\in\hS$ and
$T^{\ex}_i(G) = 0$ otherwise. Note that, for $\kappa n$ growing with
$n$, this algorithm is non-polynomial and hence cannot be used in
practice. It provides however a useful benchmark..

We have the following result showing that exhaustive search reconstructs
$S$ accurately, for any constant $\lambda$ and $\kappa$ small.
We refer to Section \ref{sec:ProofExhaustive} for a proof.
\begin{proposition}\label{propo:Exhaustive}
Let $\Psucc^{\ex} = \lim\sup_{N\to\infty}\Psucc^{(N),\ex}$ be the
asymptotic success probability of exhaustive search and assume $\kappa<1/2$. Then
\begin{align}
\Psucc^{\ex} \ge 1 - \frac{2e}{\sqrt{\kappa}}\, \exp\Big(-
\frac{\lambda (1-\kappa)b}{16\, \kappa a}\Big)\, .
\end{align}
In particular, we have the following large degree asymptotics 
as $a,b\to \infty$ with $\lambda,\kappa$ fixed
\begin{align}
\Psucc^{\ex}(b=\infty)\equiv \lim\inf_{a,b\to\infty}\Psucc^{\ex}\ge 1 - \frac{2e}{\sqrt{\kappa}}\, \exp\Big(-
\frac{\lambda (1-\kappa)}{16\, \kappa }\Big)\, ,
\end{align}
and $\Psucc^{\ex}(b=\infty)\to 1$ as $\kappa\to 0$ for any $\lambda>0$
fixed. 
\end{proposition}
\subsection{Local algorithms}
\label{sec:RigLocal}

We next give a formal definition of $t$-local algorithms. 
Let $\cG_*$ is the space of unlabeled rooted graphs, i.e. the space of
graphs with one distinguished vertex (see
--for instance-- \cite{montanari2015statistical} for more details).
Formally, an estimator $T_i(G)$ for the hidden set problem is a function
$(G,i)\mapsto T(G;i) = T_i(G)\in \{0,1\}$. Since the pair $(G,i)$ is
indeed a graph with one distinguished vertex (and the vertices labels
clearly do not matter), we can view $T$ as a
function on $\cG_*$:
\begin{align}
T: \cG_*\to\{0,1\}\,.
\end{align}
The following definition formalizes the discussion in Section
\ref{sec:TreeInterpretation} (where the definition of $\Ball_t(G,i)$
is also given). The key fact about this definition is that $t$ (the
`locality radius') is kept fixed, while the graph size can be
arbitrarily large.
\begin{definition}
Given a non-negative integer  $t$, we say that a test $T$ is
\emph{$t$-local}  if there exists a function $\cF:\cG_*\to \{0,1\}$
such that, for all $(G,i)\in\cG_*$, 
\begin{align}
T_i(G) = \cF\big(\Ball_t(G,i)\big)\, .
\end{align}
We say that a test is local, if it is $t$-local for some fixed $t$.

We denote by $\Loc(t)$ and $\Loc= \cup_{t\ge 0}\Loc(t)$ the sets of
$t$-local and  local tests.
\end{definition}

The next lemma is a well-known fact that we nevertheless state
explicitly to formalize some of the remarks of Section
\ref{sec:TreeInterpretation}.
Recall that $\Psucc^{(N)}(T)$ denotes the success probability of test
$T$, as per Eq.~(\ref{eq:PsuccT}), and  let $\Psucc(t;\frl)$ be
defined as in Eq.~(\ref{eq:PsuccFree}), with $\P_0^{(t),\fr}$, $\P_1^{(t),\fr}$, 
the laws of random variables  $\xi_0^{(t),\fr}$,
$\xi_1^{(t),\fr}$.  
\begin{lemma}
We have
\begin{align}
\sup_{T\in\Loc(t)}\lim_{N\to\infty}\Psucc^{(N)}(T) = \Psucc(t;\frl) \, .
\end{align}
In particular
\begin{align}
\sup_{T\in\Loc}\lim_{N\to\infty}\Psucc^{(N)}(T) = \Psucc(\frl) \equiv\lim_{t\to\infty} \Psucc(t;\frl)\, .
\end{align}
Further,  the maximal local success probability $\Psucc(t;\frl)$ can be achieved using
belief propagation with respect to the graphical
model  (\ref{eq:ApproxModel}) in $O(t|E|)$ time.
\end{lemma}
We will therefore valuate the fundamental limits of local algorithms
by  analyzing the quantity $\Psucc(\frl)$. The following theorem
establishes a phase transition for this quantity at $\lambda = 1/e$.
\begin{theorem}\label{thm:Main}
Consider the hidden set problem with parameters $a,b,\kappa$, and 
let $\lambda \equiv \kappa^2(a-b)^2/(1-\kappa)b$.
Then:
\begin{itemize}
\item[$(a).$] If $\lambda<1/e$, then all local algorithms have success
  probability uniformly bounded away from one. In particular, letting
$x_*(\lambda)<e$ to be the smallest positive solution of $x=
e^{\lambda x}$, we have
\begin{align}
\sup_{T\in\Loc}\lim_{N\to\infty}\Psucc^{(N)}(T) = \Psucc(\frl) \le \frac{x_*-1}{4} <
\frac{e-1}{4}\, .
\end{align}
\item[$(b).$] If $\lambda>1/e$, then local algorithms can have success
  probability arbitrarily close to one. In particular, considering the
  large degree asymptotics $a,b\to\infty$ with $\kappa,\lambda$ fixed
\begin{align}
\lim\inf_{a,b\to\infty} \Psucc(\frl)  =\Psucc^{{\rm largdeg}} (\frl;\kappa,\lambda) \, ,
\end{align}
we have
\begin{align}
\lim_{\kappa\to 0}\Psucc^{{\rm largdeg}} (\frl;\kappa,\lambda) = 1\, .
\end{align}
\end{itemize}
\end{theorem}

As a useful technical tool in proving part $(b)$ of this theorem, we establish a normal approximation
result in the spirit of Eqs.~(\ref{NormalApprox1}),
(\ref{NormalApprox2}).
In order to state this result, we recall the definition of Wasserstein
distance of order $2$, $W_2(\mu,\nu)$ between two probability measures $\mu$, $\nu$ on
$\reals$, with finite second moment $\int x^2 \nu(\de x)<\infty$,
$\int x^2 \rho(\de x)<\infty$. Namely, denoting by $\cC(\nu,\rho)$ the
family of couplings\footnote{Explicitly, $\gamma\in \cC(\nu,\rho)$ if
  it is a probability distribution on $\reals\times \reals$ such that
  $\int\gamma(A,\de y) =\nu(A)$ and $\int\gamma(\de x,A) =\rho(A)$ for
all $A$.} of $\mu$ and $\nu$, we have
\begin{align}
W_2(\nu,\rho) \equiv \left\{\inf_{\gamma\in\cC(\mu,\nu) } \int
  |x-y|^2\, 
  \gamma(\de x,\de y)\right\}^{1/2}\, .
\end{align}
Given a sequence of probability measures $\{\nu_n\}_{n\in\naturals}$
with finite second moment,
we write $\nu_n\stackrel{W_2}{\to}\nu$ if $W_2(\nu_n,\nu)\to 0$.
\begin{lemma}\label{lemma:GaussianApprox}
For $t\ge 0$, let $\xi_{0/1}^{(t),\fr}$ be the random variables
defined by the distributional recursion (\ref{eq:GeneralCavity1}),   (\ref{eq:GeneralCavity2}), 
with initial condition (\ref{eq:FreeBc}), and denote by
$\P_0^{(t),\fr}$, $\P_1^{(t),\fr}$ the corresponding laws. Further let
$\mu^{(t)}$ be defined recursively by letting $\mu^{(0)} = 0$ and
\begin{align}
\mu^{(t+1)} &= \lambda\, \F(\mu^{(t)};\kappa)\, ,\label{eq:MutRecursion}
\text{where } \F(\mu; \kappa) &= \E\left\{ \frac{1-\kappa}{\kappa +
  (1-\kappa)\exp\{-\mu/2+ \sqrt{\mu}Z\}} \right\}, \quad\quad Z\sim \normal(0,
  1).
\end{align}  
Then, considering the limit $a,b\to\infty$ with $\kappa$ fixed and
$\kappa^2(a-b)^2/((1-\kappa)b)\to\lambda\in (0,\infty)$, we have 
\begin{align}
\P_0^{(t),\fr} \stackrel{W_2}{\longrightarrow}\normal\Big(-\log\Big(\frac{1-\kappa}{\kappa}\Big)-\frac{1}{2}\,\mu^{(t)},\;
\mu^{(t)}\Big)\, ,\\
\P_1^{(t),\fr} \stackrel{W_2}{\longrightarrow}\normal\Big(-\log\Big(\frac{1-\kappa}{\kappa}\Big)+\frac{1}{2}\,\mu^{(t)},\;
\mu^{(t)}\Big)\, .
\end{align}
\end{lemma}
The proof of this lemma  is presented in Section \ref{sec:ProofGaussian}.
%
%
\section{Discussion and related work}
\label{sec:Related}

As mentioned in the introduction, the problem of identifying a highly
connected subgraph in an otherwise random graph has been studied
across multiple communities. 
Within statistical theory, Arias-Castro and Verzelen
\cite{arias2014community,verzelen2013community}
established necessary and sufficient conditions for distinguishing 
a purely random graph, from one with a hidden community.  
With the scaling adopted in our paper, this `hypothesis testing' problem requires
to distinguish between the following two hypotheses:
\begin{align*}
H_0:& \;\; \mbox{Each edge is present independently with probability
      $b/N$},\\
H_1:& \;\; \mbox{Edges within the community are present with
      probability $a/N$.}\\
&\;\; \mbox{ Other edges are present with probability $b/N$}.
\end{align*}
Note that this problem is trivial in the present regime and can be
solved --for instance-- by counting the number of edges in $G$.

The sparse graph regime studied in the present paper was also recently considered 
in a series of papers that analyzes community detection problems using
ideas from statistical physics
\cite{decelle2011inference,decelle2011asymptotic,krzakala2013spectral}.
 The focus of these papers is
on a setting whereby the graph $G$ contains $k\ge 2$ non-overlapping communities, each
of equal size $N/k$. Using our notation, vertices within the
same community are connected with probability $a/N$ and vertices
belonging to different communities are connected with probability
$b/N$.
Interestingly, the results of  \cite{decelle2011asymptotic} point at a similar phenomenon as the one studied here
for $k\ge 5$. Namely, for a range of
parameters
the community structure can be identified by exhaustive search, but
low complexity algorithms appear to fail. 

Let us mention that the very same phase transition structure arises in
other inference problem, for instance in decoding sparse graph error
correcting codes, or solving planted constraint satisfaction problems
\cite{RiU08,MezardMontanari,achlioptas2006solution,zdeborova2011quiet}.
 A unified formalism for all of these problems is adopted in \cite{abbe2013conditional}.
All of these problems  present a regime of model parameters whereby 
a large gap separates the optimal estimation accuracy, from the
optimal accuracy achieved by known polynomial time algorithms.
Establishing that such a gap cannot be closed under standard
complexity-theoretic assumptions is an outstanding challenge.
(See \cite{hajek2014computational} for partial evidence in this
direction
--albeit in a different regime.)
One can nevertheless gain useful insight by studying classes of
algorithms with increasing sophistication.
\begin{description}
\item[Local algorithms] are a natural starting point for sparse graph
  problems.
The problem of  finding a large independent
set in a sparse random graph is closely related to the one studied here. Indeed an independent set can be
viewed as a subset of vertices that is `less-connected' than the
background (indeed is a subset of vertices such that the induced
subgraph has no edge). 

The largest independent set in a uniformly random regular graph
with $N$ vertices of
degree $d$ has typical size $\alpha(d)\, N +o(N)$ where, for large
bounded degree $d$,
$\alpha(d) = 2d^{-1}\log d (1+o_d(1))$. Hatami,  Lov\'asz  and Szegedy
\cite{hatami2012limits} conjectured that local algorithms can
find independent sets of almost maximum size --up to sublinear
terms in $N$. Gamarnik and Sudan \cite{gamarnik2014limits} recently
disproved this conjectured and demonstrated a constant multiplicative
gap for 
local algorithms. Roughly speaking, for large degrees no local algorithm can produce an
independent set of size larger than $86\%$ of the optimum. This factor of
$86\%$ was later
improved by Rahman and Virag \cite{rahman2014local} to $50\%$. 
This gap is analogous to the gap in estimation error established in
the present paper.
We refer to \cite{gamarnik2014local} for a broader review of this line
of work.

As mentioned before, belief propagation (when run for an arbitrary
\emph{fixed} number of iterations) is a special type of local
algorithm. Further it is basically optimal (among local algorithms) for Bayes estimation on
locally tree like graphs. The gap between belief propagation decoding
and optimal decoding is well studied in the context of coding \cite{RiU08,MezardMontanari}.
\item[Spectral algorithms.] Let $A_N$ be the adjacency matrix of the graph
  $G_N$ (for simplicity we set $(A_N)_{ii}\sim\Bernoulli(a/N)$ for
  $i\in S$, and $(A_N)_{ii}\sim\Bernoulli(b/N)$ for $i\not\in S$). We
  then have
\begin{align}
\E\{A_N|S\} = \frac{a-b}{n} \, \bone_S\bone_S^{\sT} + \frac{b}{n} \,
  \bone\bone^{\sT}\, .
\end{align}
This suggests that the principal eigenvector of
$(A_N-(b/n)\bone\bone^{\sT})$ should be localized on the set $S$.
Indeed this approach succeeds in the dense case (degree of order $n$), allowing to
reconstruct $S$ with high probability \cite{alon1998finding}.

In the sparse graph setting considered here, the approach fails
because the operator norm $\|A_N-\E\{A_N|S\}\|_2$ is unbounded as
$N\to\infty$.
Concretely, the sparse graph $G_N$ has large eigenvalues of order
$\sqrt{\log N/\log\log N}$ localized on the vertices of largest
degree. This point was already discussed in several related problems
\cite{feige2005spectral,coja2010graph,keshavan2010matrix,krzakala2013spectral,mossel2013proof}.
Several techniques have been proposed to address this problem, the
crudest one being to remove high-degree vertices.

We do not expect spectral techniques  to overcome the limitations
of local algorithms in the present problem, even in their advanced
forms that take into account degree heterogeneity.  Evidence for this
claim is provided by studying the dense graph case, in which degree
heterogeneity does not pose problems. In that case
 spectral techniques are known to fail for $\lambda<1$
 \cite{deshpande2013finding,montanari2014limitation}, and hence are 
strictly inferior to (local) message passing algorithms that
succeed\footnote{Note that the definition of $\lambda$ in the present
  paper correspond to $\lambda^2$ in \cite{deshpande2013finding,montanari2014limitation}.}
for any $\lambda>1/e$. 
\item[Semidefinite relaxations.] Convex relaxations provide a natural
class of polynomial time algorithms that are more powerful than
spectral approaches. Feige and Krauthgamer \cite{feige2000finding,feige2003probable}
studied the Lov\'asz-Schrijver hierarchy of semidefinite programming
(SDP) relaxations for the  hidden clique problem. In that setting,
each round of the hierarchy yields a constant factor improvement in
clique size, at the price of increasing complexity. It would be
interesting to extend their analysis to the sparse regime. It is
unclear whether SDP hierarchies are more powerful than simple local
algorithms in this case. 
\end{description}

Let us finally mention that the probability measure
(\ref{eq:ApproxModel})
can be interpreted as the Boltzmann distribution for a system of 
$\kappa N$ particles on the graph $G$, with fugacity $\gamma$, and
interacting attractively (for $\rho>1$). Statistical mechanics
analogies were previously exploited in \cite{iovanella2007some,gaudilliere2011phase}.
(See also \cite{hu2012phase} for the general community detection problem.)

\section*{Acknowledgements}

I am grateful to Yash Deshpande for carefully reading this manuscript
and providing valuable feedback.
This work was partially supported by the NSF grants CCF-1319979 and
DMS-1106627, and the
grant AFOSR FA9550-13-1-0036. 
%
%
\appendix

\section{Proof of Proposition \ref{propo:Exhaustive}}
\label{sec:ProofExhaustive}

For the sake of simplicity, we shall assume a slightly modified model
whereby the hidden set $S$ is uniformly random with size $|S|=k$, with $k/N\to\kappa$.
Recall that, under the independent model (\ref{eq:KappaDef}) $|S|\sim
  \Binom(n,\kappa)$ and hence is tightly concentrated  around its mean
  $\kappa n$. Hence, the result the independent model follows by a
  simple conditioning argument. 

Let $L \equiv |\hS\cap S|$. By exchangeability of the graph
vertices, we have
\begin{align}
\Psucc^{(N),\ex} &=  \prob\big(T_i(G) = 1\big| i\in S\big) + \prob\big(T_i(G) =
0\big|i\not\in S\big)-1 \\
&= \E\Big\{\frac{L}{k}+ \frac{N-2k+L}{N-k}-1\Big\}\\
& = \E\Big\{\frac{L}{k}- \frac{k-L}{N-k}\Big\}\ge 1 - 2\E\Big\{\frac{k-L}{k}\Big\}\, ,\label{eq:ProofPropo-Simple}
\end{align}
where the last inequality follows since, without loss of generality, $N-k>k$.
Setting $x_*\equiv  (e/\sqrt{\kappa})\exp\big(-\lambda(1-\kappa)b/(16\,
\kappa  a)\big)$, we will prove that for any $\delta> 0$ there exists
$c(\delta)>0$ such that
\begin{align}
\prob\big(L\le k (1-x_*-\delta)\big)\le 2\,  e^{-n\,c(\delta)}\, . \label{eq:BasicProofPropo}
\end{align}
The claim the follows by using the inequality
(\ref{eq:ProofPropo-Simple}) together with the
fact that $(k-L)/k\le 1$. 

For two sets $A$, $B\subseteq V= [N]$, we let $E(A,B)$ the number of
edges $(i,j)\in E$ such that $\{i,j\}\subseteq A$, but
$\{i,j\}\not\subseteq B$. 
In order to prove Eq.~(\ref{eq:BasicProofPropo}) note that, for
$\ell\in\{0,1,\dots,k\}$
\begin{align}
\prob(L = \ell)&\le \prob\big(\exists R\subseteq V:\; |R|=k,\,|R\cap
S|=\ell,\, E(R,S)\ge E(S,R)\big)\, . \label{eq:ToBeExplained}
\end{align}
To see this notice that, by definition, if $L=\ell$ then $|\hS\cap
S|=\ell$. This mean that there must exists at least one set
$R\subseteq [n]$ satisfying the following conditions:
\begin{itemize}
\item $|R| = k$.
\item $|R\cap S|=\ell$.
\item $E(R)\ge E(S)$.
\end{itemize}
Indeed $\hS$ is such a set. This immediately implies
Eq.~(\ref{eq:ToBeExplained})
by noticing that $E(S,R) = E(S)-E(S\cap R)$ and $E(R,S) = E(R)-E(S\cap R)$.
By a union bound (setting $m\equiv \binom{k}{2}-\binom{\ell}{2}$):
\begin{align}
\prob(L = \ell)& \le \sum_{j=0}^m \prob\big(\exists R_1\subseteq S,
R_2\subseteq V\setminus S:\; |R_1|=\ell,\,  |R_2|=k-\ell,\,
E(S,R_1)\le j,\,
E(R_1\cup R_2,S)\le j\big)\\
& \le \sum_{j=0}^m \binom{k}{\ell}\binom{N-k}{k-\ell}\,
\prob\big(\Binom(m;a/N)\le j\big)
\,\,\prob\big(\Binom(m;b/N)\ge j\big)\, .
\end{align}
In the last inequality we used union bound and the fact that edges
contributing to $E(S,R_1)$ and $E(R_1\cup R_2,S)$ are independent.
Using Chernoff bound on the tail of binomial random variables
(with $D(q||p) = q\log(q/p)+(1-q)\log((1-q)/(1-p))$ the
Kullback-Leibler divergence between two Bernoulli random variables),
we get
\begin{align}
\prob(L = \ell) & \le (m+1) \binom{k}{\ell}\binom{N-k}{k-\ell}\,
\max_{j\in [bm/n,am/n]\cap \naturals} 
\prob\big(\Binom(m;a/N)\le j\big)
\,\,\prob\big(\Binom(m;b/N)\ge j\big)\\
&\le (m+1)\binom{k}{\ell}\binom{N-k}{k-\ell}\,
\exp\Big\{-m\, \min_{j\in [bm/n,am/n]} \big[D(j/m|| a/N)+D(j/m||
b/N)\big]\Big\}, . \label{eq:BoundWKL}
\end{align}
Here, the first inequality follows because both probabilities are
increasing for $j<bm/N$ and decreasing for $j>am/N$.
We further note that,
$\frac{\de^2 D(x||p)}{\de x^2}\ge 1+x^{-1}$ and therefore,  for $q, p\in [0,1]$,
\begin{align}
D(q||p) \ge \frac{1}{2}\Big(\frac{1}{\max(p,q)}+1\Big)(q-p)^2\, .
\end{align}
This implies that, for $p_1<p_2$, we have
\begin{align}
\min_{x\in [p_1,p_2]} \big[D(x||p_1)+D(x||p_2)\big]&\ge
\frac{1}{2}\Big(\frac{1}{p_2}+1\Big)\,
\min_{x\in [p_1,p_2]} \big[(x-p_1)^2+(x-p_2)^2\big]\\
&\ge
\frac{1}{4}\Big(\frac{1}{p_2}+1\Big)\, (p_1-p_2)^2\, .
\end{align}
We substitute the last inequality in Eq.~(\ref{eq:BoundWKL}), 
together with the bounds $\binom{a}{b}\le \min[(e a/b)^b,( ea/(a-b))^{a-b}]$
\begin{align}
\prob(L = \ell) &\le  (m+1) \Big(\frac{ke}{k-\ell}\Big)^{k-\ell}
\Big(\frac{Ne}{k-\ell}\Big)^{k-\ell}\exp\Big\{-\frac{m}{4}\Big(1+\frac{N}{a}\Big)\Big(\frac{a}{N}-\frac{b}{N}\Big)^2\Big\}\, .
\end{align}
We let $\ell = k (1-x) =\kappa N(1-x)$ whence
\begin{align}
m = \binom{k}{2}-\binom{\ell}{2}\;\ge\; \frac{k}{2}(k-\ell) =
\frac{N^2\kappa^2}{2}x\, .
\end{align}
We therefore get
\begin{align}
\prob(L = \ell) &\le  (m+1)
\Big(\frac{e}{x\sqrt{\kappa}}\Big)^{2\kappa N x}\,
\exp\Big\{-\frac{Nx}{8}\, \frac{\kappa^2(a-b)^2}{a}\Big\}\\
& \le (m+1) \left\{\frac{e}{x\sqrt{\kappa}}\,
\exp\Big(-\frac{\lambda(1-\kappa)b}{16\kappa\, a}\Big)\right\}^{2\kappa
  Nx}\, .
\end{align}
For $x\ge x_*+\delta$, the argument in parenthesis is smaller than
$e^{-c(\delta)/(2\kappa x)}$ and therefore 
\begin{align}
\prob(L = \ell) &\le (m+1) \, e^{-Nc(\delta)},.
\end{align}
Summing over $\ell\le k(1-x_*-\delta)$, we get 
$\prob(L k(1-x_*-\delta)) \le k(m+1) \, e^{-Nc(\delta)}$ which
implies  the claim (\ref{eq:BasicProofPropo}), after eventually
adjusting $c(\delta)$, since $k(m+1)\le N^3$. 

\section{Proof of Theorem \ref{thm:Main}}
\label{sec:ProofLocal}

\subsection{Proof of Lemma \ref{lemma:GaussianApprox}}
\label{sec:ProofGaussian}

Throughout this section we will drop the superscript $\frl$ from
$\xi_{0/1}^{(t),\fr}$ and $\P_{0/1}^{(t)}$.

Recall that convergence in $W_2$ distance is
equivalent to weak convergence, plus convergence of the first two
moments \cite[Theorem 6.9]{villani2008optimal}.
We will prove by the following by induction over $t$:
\begin{enumerate}
\item[I.] The first moments $\E\{|\xi_0^{(t)}|\}$, $\E\{|\xi_1^{(t)}|\}$
  are finite and we have
\begin{align}
\lim_{a,b\to\infty}\E\{\xi_0^{(t)}\} &=
-\log\Big(\frac{1-\kappa}{\kappa}\Big)-\frac{1}{2}\,\mu^{(t)}\, ,\label{eq:LimitExpXi0}\\
\lim_{a,b\to\infty}\E\{\xi_1^{(t)}\} &=
-\log\Big(\frac{1-\kappa}{\kappa}\Big)+\frac{1}{2}\,\mu^{(t)}\, .
\end{align}
\item[II.] The variances $\Var(\xi_0^{(t)})$, $\Var(\xi_1^{(t)})$ are
  finite and they converge
\begin{align}
\lim_{a,b\to\infty}\Var(\xi_0^{(t)}) &=\mu^{(t)}\, ,\label{eq:LimitVarXi0}\\
\lim_{a,b\to\infty}\Var (\xi_1^{(t)}) &=\mu^{(t)}\, .
\end{align}
\item[III.] Weak convergence
\begin{align}
\P_0^{(t)}\Rightarrow\normal\Big(-\log\Big(\frac{1-\kappa}{\kappa}\Big)-\frac{1}{2}\,\mu^{(t)},\;
\mu^{(t)}\Big)\, ,\label{eq:LimitWeakP0}\\
\P_1^{(t)} \Rightarrow\normal\Big(-\log\Big(\frac{1-\kappa}{\kappa}\Big)+\frac{1}{2}\,\mu^{(t)},\;
\mu^{(t)}\Big)\, .
\end{align}
\end{enumerate}
These claims  obviously hold for
$t=0$. Next assuming that they hold up to iteration $t$, we need to
prove them for iteration $t+1$. For the sake of brevity, we will only
present this calculation for $\P_0^{(t+1)}$, since the derivation for
$\P_1^{(t+1)}$ is completely analogous. 

\vspace{0.3cm}

Let us start by considering Eq.~(\ref{eq:LimitExpXi0}).
First notice that the absolute value of right-hand side of Eq.~(\ref{eq:GeneralCavity1})
is upped bounded by
\begin{align}
h + C_2\sum_{i=1}^{L_{00}} (1+|\xi^{(t)}_{0,i}|)
+C_2\sum_{i=1}^{L_{01}} (1+|\xi^{(t)}_{1,i}|) \, ,\label{eq:UpperBoundRHS}
\end{align}
and hence  $\E|\xi^{(t+1)}_0|<\infty$  follows from the induction
hypothesis I$(t)$ and the fact that $L_{00}, L_{01}$ are Poisson.
Next to prove Eq.~(\ref{eq:LimitExpXi0}), we take expectation of Eq.~(\ref{eq:GeneralCavity1}),  and
let, for simplicity, $l(\kappa)\equiv \log((1-\kappa)/\kappa)$:
\begin{align}
\E\{\xi_0^{(t+1)}\} &=-l(\kappa)-\kappa(a-b)+
(1-\kappa)b \,
\E\log\left(1+(\rho-1)\frac{e^{\xi_0^{(t)}}}{1+e^{\xi_0^{(t)}}}\right)\\
&\phantom{=}+\kappa b  \,
\E\log\left(1+(\rho-1)\frac{e^{\xi_1^{(t)}}}
  {1+e^{\xi_1^{(t)}}}\right)\nonumber\\
& = -l(\kappa)-\kappa(a-b)+ \label{eq:ExiExpansion}\\
&\phantom{=}+(1-\kappa)(a-b)
\E\left(\frac{e^{\xi_0^{(t)}}}{1+e^{\xi_0^{(t)}}}\right) +\kappa (a-b) \E\left(\frac{e^{\xi_1^{(t)}}}{1+e^{\xi_1^{(t)}}}\right) \nonumber\\
&\phantom{=}-(1-\kappa) \frac{(a-b)^2}{2b}
\E\left\{\left(\frac{e^{\xi_0^{(t)}}}{1+e^{\xi_0^{(t)}}}\right)^2\right\}
-\kappa
\frac{(a-b)^2}{2b}\E\left\{\left(\frac{e^{\xi_1^{(t)}}}{1+e^{\xi_1^{(t)}}}\right)^2\right\}
+O\left(\frac{(a-b)^3}{b^2}\right)\nonumber
\end{align}
where the last equality follows from bounded convergence, since, for
all $x\in\reals$, $0\le e^x/(1+e^x)\le 1$.
Note that the laws of  $\xi_0^{(t)}$ and $\xi_1^{(t)}$ satisfy the
symmetry property (\ref{eq:Symmetry}). Hence, for any measurable
function $g:\reals\to\reals$  such that the expectations below make
sense, we have
\begin{align}
(1-\kappa) \E\, g(\xi_0^{(t)}) + \kappa \E\, g(\xi_1^{(t)}) =\kappa
\E\{(1+e^{-\xi_1^{(t)}} )\, g(\xi_1^{(t)}) \}\, .
\end{align}
In particular applying this identity to $g(x) = e^x/(1+e^x)$ and $g(x)
= [e^x/(1+e^x)]^2$, we get
\begin{align}
(1-\kappa)
\E\left(\frac{e^{\xi_0^{(t)}}}{1+e^{\xi_0^{(t)}}}\right) +\kappa\,
\E\left(\frac{e^{\xi_1^{(t)}}}{1+e^{\xi_1^{(t)}}}\right) &=\kappa\, ,\\
(1-\kappa)
\E\left\{\left(\frac{e^{\xi_0^{(t)}}}{1+e^{\xi_0^{(t)}}}\right)^2\right\}+\kappa\,
\E\left\{\left(\frac{e^{\xi_1^{(t)}}}{1+e^{\xi_1^{(t)}}}\right)^2\right\}
&=\kappa \E\left(\frac{e^{\xi_1^{(t)}}}{1+e^{\xi_1^{(t)}}}\right)\, .
\label{eq:SecondIdentity}
\end{align}
Substituting in Eq.~(\ref{eq:ExiExpansion}), and expressing $a$ in
terms of $b,\kappa,\lambda$ we get
\begin{align}
\E\{\xi_0^{(t+1)}\} &=-l(\kappa) - \frac{(1-\kappa)\lambda}{2\kappa}\, 
\E\left(\frac{e^{\xi_1^{(t)}}}{1+e^{\xi_1^{(t)}}}\right)
+O(b^{-1/2})\\
 &=-l(\kappa) - \frac{(1-\kappa)\lambda}{2\kappa}
\E\left(\frac{1}{1+\exp\big\{ l(\kappa)-\mu^{(t)}/2+\sqrt{\mu^{(t)}} \, Z\}}\right)
+o_b(1)\, ,
\end{align}
where $o_b(1)$ denotes a quantity vanishing as $b\to\infty$.
The last equality follows from induction hypothesis III$(t)$
and the fact that $g(x) = 1/(1+e^{-x})$ is bounded continuous,
with $Z\sim\normal(0,1)$. This yields the desired claim
(\ref{eq:LimitExpXi0}) after comparing with
Eq.~(\ref{eq:MutRecursion}).

\vspace{0.3cm}

Consider next Eq.~(\ref{eq:LimitVarXi0}). The upper bound on the
right-hand side of Eq.~(\ref{eq:GeneralCavity1}) given by
Eq.~(\ref{eq:UpperBoundRHS}) immediately imply that
$\Var(\xi^{(t+1)}_0)<\infty$.  In order to
estabilish Eq.~(\ref{eq:LimitVarXi0}), we recall an elementary
formula for the variance of a Poisson sum.  If $L$ is a
Poisson random variable and $\{X_i\}_{i\ge 1}$ are i.i.d. with finite
second moment, then 
\begin{align}
\Var\Big(\sum_{i=1}^LX_i\Big) = \E(L) \, E(X_1^2)\, .
\end{align}
Applying this to Eq.~(\ref{eq:GeneralCavity1}), and expanding for
large $b$ thanks to the bounded convergence theorem, we get
\begin{align}
\Var(\xi_0^{(t+1)}) &=
(1-\kappa)b \,
\E\left\{\big[\log\left(1+(\rho-1)\frac{e^{\xi_0^{(t)}}}{1+e^{\xi_0^{(t)}}}\right)\big]^2\right\}+\kappa b  \,
\E\left\{\big[\log\left(1+(\rho-1)\frac{e^{\xi_1^{(t)}}}
{1+e^{\xi_1^{(t)}}}\right)\big]^2\right\}\\
&= (1-\kappa) \frac{(a-b)^2}{b}
\E\left\{\left(\frac{e^{\xi_0^{(t)}}}{1+e^{\xi_0^{(t)}}}\right)^2\right\}
+\kappa\, 
\frac{(a-b)^2}{b}\E\left\{\left(\frac{e^{\xi_1^{(t)}}}{1+e^{\xi_1^{(t)}}}\right)^2\right\}
+O(b^{-1/2})\\
& = \kappa 
\frac{(a-b)^2}{b}\E\left(\frac{e^{\xi_1^{(t)}}}{1+e^{\xi_1^{(t)}}}\right)
+O(b^{-1/2})\, ,
\end{align}
where the last equality follows by applying again
Eq.~(\ref{eq:SecondIdentity}). By using the induction hypothesis
III$(b)$ and the fact that $g(x) = (1+e^{-x})$ is bounded Lipschitz,
\begin{align}
\lim_{a,b\to\infty}\Var(\xi_0^{(t+1)}) &=\frac{1-\kappa}{\kappa}\,
\lambda\, 
\E\left(\frac{1}{1+\exp\big\{ l(\kappa)-\mu^{(t)}/2+\sqrt{\mu^{(t)}}
    \, Z\}}\right) = \lambda \, \F(\mu^{(t)};\kappa)\, ,
\end{align}
which is Eq.~(\ref{eq:MutRecursion}). 

\vspace{0.3cm}

We finally consider Eq.~(\ref{eq:LimitWeakP0}). By subtracting the
mean, we can rewrite Eq.~(\ref{eq:GeneralCavity1}) as
\begin{align}
\xi_0^{(t+1)}- \E\{\xi_0^{(t+1)}\}&\ed \sum_{i=1}^{L_{00}} X_i+
\sum_{i=1}^{L_{01}} Y_i +(L_{00}-\E L_{00}) \E f(\xi^{(t)}_{0,1})
+(L_{01}-\E L_{01}) \E f(\xi^{(t)}_{1,1})
\, ,
\end{align}
where $X_i = f(\xi^{(t)}_{0,i})-\E f(\xi^{(t)}_{0,i})$,
$Y_i = f(\xi^{(t)}_{1,i})-\E f(\xi^{(t)}_{1,i})$. 
Note that $X_i$, $Y_i$ have zero mean and, by the calculation above,
they have variance $\E\{X_i^2\} = \E\{Y_i^2\} = O(1/b)$.
Denoting the right hand side by $S_b$:
\begin{align}
S_b = \sum_{i=1}^{\E L_{00}} X_i+
\sum_{i=1}^{\E L_{01}} Y_i +(L_{00}-\E L_{00})  \E f(\xi^{(t)}_{0,1})
+(L_{01}-\E L_{01}) \E f(\xi^{(t)}_{1,1}) +o_P(1) \label{eq:Sb}
\, ,
\end{align}
because (for instance) $\sum_{i=1}^{L_{00}} X_i-\sum_{i=1}^{\E L_{00}}
X_i$ is a sum of order $\sqrt{b}$ independent random variables with
zero mean and variance of order $1/b$.
Note that
\begin{align}
\lim_{a,b\to\infty} \E L_{0,0}\Var(X_1) + 
\lim_{a,b\to\infty} \E L_{0,1}\Var(Y_1) +
\lim_{a,b\to\infty} \Var(L_{0,0}) \E f(\xi^{(t)}_{0,1}) +
\lim_{a,b\to\infty} \Var(L_{0,1})\E f(\xi^{(t)}_{1,1}) \nonumber\\
=
\lim_{a,b\to\infty} \big\{(1-\kappa)b\E[ f(\xi^{(t)}_{0,1})^2 ] +
\kappa b\E[ f(\xi^{(t)}_{1,1})^2 ] \big\} = \mu^{(t+1)}\, ,
\end{align}
where the last equality follows by the calculation above.
Hence, by applying the central limit theorem to each of the four terms in 
Eq.~(\ref{eq:Sb}) and noting that they are independent, 
we conclude that  $S_{b}$ converges in distribution to  $\normal(0,\mu^{(t+1)})$.

\subsection{Proof of Theorem \ref{thm:Main}.$(a)$}

Define the event $A=\{\xi\ge \log(\kappa/(1-\kappa))\}$,
and write $\P_{0/1}^{(t)}$ for $\P_{0/1}^{(t),\fr}$.
From Eq.~(\ref{eq:PsuccFree}) we have
\begin{align}
\Psucc(t;\frl) &=\P_0^{(t)}(A^c) +\P_1^{(t)}(A)-1\\
& =  \P_1^{(t)}(A)
                 -\P_0^{(t)}(A) \\
& = \Er_0^{(t)}\Big\{\ind_A\, \frac{\de \P^{(t)}_1}{\de \P^{(t)}_0}\Big\}
  -\P_0^{(t)}(A) \\
& \le \Er_0^{(t)}\left\{\Big(\frac{\de \P^{(t)}_1}{\de \P^{(t)}_0}-1\Big)^{2}\right\}^{1/2}\, \P_0^{(t)}(A)^{1/2}
  -\P_0^{(t)}(A) \\
& \le \sup_{q\ge 0}\left\{
\Er_0^{(t)}\left\{\Big(\frac{\de \P^{(t)}_1}{\de \P^{(t)}_0}-1\Big)^{2}\right\}^{1/2}\, q
  -q^2\right\}\\
& =\frac{1}{4}\Er_0^{(t)}\left\{\Big(\frac{\de \P^{(t)}_1}{\de
  \P^{(t)}_0}-1\Big)^{2}\right\}\, .
\end{align}
Using  Eq.~(\ref{eq:Symmetry}), and the fact that
$\Er_0^{(t)}\Big(\frac{\de \P^{(t)}_1}{\de \P^{(t)}_0}\Big)=1$, we get
\begin{align}
\Psucc(t;\frl) \le
                 \frac{1}{4}\left(\Big(\frac{1-\kappa}{\kappa}\Big)^2
\E\{e^{2\xi_0^{(t)}}\}-1\right)\, . \label{eq:BasicTVBound}
\end{align}
Call $x_t\equiv
(1-\kappa)^2\kappa^{-2}\E\{e^{2\xi_0^{(t)}}\}$. By
the initialization (\ref{eq:FreeBc}), $x_0 = 1$. 
Taking exponential
moments of Eq.~(\ref{eq:GeneralCavity1}), we get
\begin{align}
x_{t+1} = 
\exp\left\{-2\kappa a+ (2\kappa-1)b
  +(1-\kappa)b\,\E\left[\left(\frac{1+\rho\, e^{\xi_0^{(t)}} }{
  1+ e^{\xi_0^{(t)}} }\right)^2\right]
+\kappa b \,\E\left[\left(\frac{1+\rho\, e^{\xi_1^{(t)}} }{
  1+ e^{\xi_1^{(t)}} }\right)^2\right]
 \right\}\, .
\end{align}
Note that by Eq.~(\ref{eq:Symmetry}), for any measurable
function $g:\reals\to\reals$  such that the expectations below make
sense, we have
\begin{align}
(1-\kappa) \E\, g(\xi_0^{(t)}) + \kappa \E\, g(\xi_1^{(t)}) =(1-\kappa)\,
\E\{(1+e^{\xi_0^{(t)}} )\, g(\xi_0^{(t)}) \}\, .
\end{align}
Applying this to $g(x) = (1+\rho e^x)^2/(1+e^x)^2$, we get
\begin{align}
x_{t+1} = 
\exp\left\{-2\kappa a+ (2\kappa-1)b
  +(1-\kappa)b\,\E\left[\frac{(1+\rho\, e^{\xi_0^{(t)}} )^2}{
  1+ e^{\xi_0^{(t)}} }\right]
 \right\}\, .
\end{align}
Now we claim that, for  $z\ge 0$, we have
\begin{align}
\frac{(1+\rho z)^2}{1+z}\le 1+(2\rho-1)z+(\rho-1)^2z^2\, .
\end{align}
This can be checked, for instance, by multiplying both sides by
$(1+z)$ and simplifying. Using $\E\{e^{\xi_0^{(t)}}\} =
\kappa/(1-\kappa)$ and $\E\{e^{2\xi_0^{(t)}}\} =
\kappa^2x_t/(1-\kappa)^2$, we get
\begin{align}
x_{t+1} &\le
\exp\left\{-2\kappa a+ (2\kappa-1)b
  +(1-\kappa)b\,\Big(1+(2\rho-1) \frac{\kappa}{1-\kappa} +(\rho-1)^2
\Big(\frac{\kappa}{1-\kappa}\Big)^2 x_t \Big)
 \right\}\\
& = e^{\lambda\, x_t}\, .
\end{align}
Let $\ox_t$ be the solution of the above recursion with equality,
i.e. $\ox_0 = 1$ and
\begin{align}
\ox_{t+1} = e^{\lambda\ox_t}\, .
\end{align}
It is a straightforward exercise to see that $\ox_t$ is monotone
increasing in $t$ and $\lambda$. Further, for $\lambda\le 1/e$, $\lim_{t\to\infty}\ox_t(\lambda)
= x_*(\lambda)$ the smallest positive solution of $x= e^{\lambda x}$,
and $x_*(\lambda)\le x_*(1/e) = e$.
Hence $x_t\le \ox_t\le x_*(\lambda)$ which, together with
Eq.~(\ref{eq:BasicTVBound}) finishes the proof.

\subsection{Proof of Theorem \ref{thm:Main}.$(b)$}

Note that by monotonicity $\Psucc(\frl)\ge \Psucc(t;\frl)$, and hence 
it is sufficient to lower bound the limit of the latter quantity.
By  Lemma \ref{lemma:GaussianApprox}, we have
\begin{align}
\lim_{a,b\to\infty}\Psucc(t;\frl) = 1-2\,
  \Phi\Big(-\sqrt{\mu^{(t)}}/2\Big)\, ,
\end{align}
where $\Phi(x) \equiv\int_{-\infty}^x e^{-z^2/2}\de z/\sqrt{2\pi}$ is
the Gaussian distribution, and $\mu^{(t)}$ is defined recursively by 
Eq.~(\ref{eq:MutRecursion}) with $\mu^{(0)} = 0$. 
Hence  for all $t\ge 0$
\begin{align}
\lim_{\kappa\to 0}\Psucc^{{\rm largdeg}} (\frl;\kappa,\lambda) \ge
  \lim_{\kappa\to 0}\Big\{1-2\,
  \Phi\Big(-\sqrt{\mu^{(t)}}/2\Big)\Big\}\, .
\end{align}
It is therefore sufficient to prove that
\begin{align}
\lim_{t\to\infty}\lim_{\kappa\to 0}\mu^{(t)} = \infty\, .
\end{align}
Now by monotone convergence, we have
\begin{align}
\lim_{\kappa\to 0}\F(\mu;\kappa) = \E\{e^{(\mu/2)-\sqrt{\mu} Z}\} =
  e^{\mu}\, .
\end{align}
Further $\F(\mu;\kappa)$ increases monotonically towards its limit as
$\kappa\to 0$. Furthermore, $\F(\mu; \kappa)$ is increasing in $\mu$ for
any fixed $\kappa\ge 0$. 
By induction over $t$ we prove that $\lim_{\kappa\to
  0}\mu^{(t)} = \omu^{(t)}$ (the limit being monotone from below), where $\omu^{(0)} = 0$ and for all $t\ge 0$
\begin{align}
\omu^{(t+1)} = \lambda \, e^{\omu^{(t)}}\, .\label{eq:OmuDef}
\end{align}
In order to prove this claim, note that the base case of the induction
is trivial and (writing explicitly the dependence on $\kappa$
\begin{align}
\mu^{(t+1)}(\kappa)\le \lambda \, e^{\mu^{(t)}(\kappa)}\le \lambda \,
  e^{\omu^{(t)}} \equiv\omu^{(t+1)}\, .
\end{align}
On the other hand for a fixed $\kappa_0>0$ 
\begin{align}
\lim_{\kappa\to 0}\mu^{(t+1)}(\kappa)\ge \lambda\lim_{\kappa\to
  0}\F(\mu^{(t)}(\kappa_0);\kappa) = \lambda\,
  e^{\mu^{(t)}(\kappa_0)}\,.
\end{align}
The claim follows since $\kappa_0$ can be taken arbitrarily small. 

Now it is easy to show from Eq.~(\ref{eq:OmuDef})  that
$\lim_{t\to\infty}\omu^{(t)}=\infty$ for
$\lambda>1/e$ (this is is indeed closely related to the sequence
$\ox_t$ constructed in the previous section, since $\ox_t = \exp(\omu^{(t)})$).

\bibliographystyle{amsalpha}

\newcommand{\etalchar}[1]{$^{#1}$}
\providecommand{\bysame}{\leavevmode\hbox to3em{\hrulefill}\thinspace}
\providecommand{\MR}{\relax\ifhmode\unskip\space\fi MR }
\providecommand{\MRhref}[2]{%
  \href{http://www.ams.org/mathscinet-getitem?mr=#1}{#2}
}
\providecommand{\href}[2]{#2}

\end{document}